\documentclass[twoside]{article}

\usepackage[accepted]{aistats2025}

\usepackage{graphicx}%
\usepackage{multirow}%
\usepackage{amsmath,amssymb,amsfonts}%
\usepackage{amsthm}%
\usepackage{mathrsfs}%
\usepackage[title]{appendix}%
\usepackage{xcolor}%
\usepackage{textcomp}%
\usepackage{manyfoot}%
\usepackage{booktabs}%
\usepackage{algorithm}%
\usepackage{algorithmicx}%
\usepackage{algpseudocode}%
\usepackage{listings}%
\usepackage{hyperref}

\begin{document}

%

%

\twocolumn[

\aistatstitle{Inference-based GAN Video Generation}

\aistatsauthor{Jingbo Yang \And Adrian G. Bors }

\aistatsaddress{\\ Department of Computer Science, University of York, York, YO10 5GH, UK\\
E-mail: jingbo\_yang2019@163.com, adrian.bors@york.ac.uk} ]

\begin{abstract}
   Video generation has seen remarkable progress thanks to advancements in generative deep learning. However, generating long sequences remains a significant challenge.
   Generated videos should not only display coherent and continuous movement but also meaningful movement in successions of scenes. Models such as GANs, VAEs, and Diffusion Networks have been used for generating short video sequences, typically up to 16 frames. In this paper, we first propose a new type of video generator by enabling adversarial-based unconditional video generators with a variational encoder, akin to a VAE-GAN hybrid structure. The proposed model, as in other video deep learning-based processing frameworks, incorporates two processing branches, one for content and another for movement. However, existing models struggle with the temporal scaling of the generated videos. 
   Classical approaches often result in degraded video quality when attempting to increase the generated video length, especially for significantly long sequences. To overcome this limitation, our research study extends the initially proposed VAE-GAN video generation model by employing a novel, memory-efficient approach to generate long videos composed of hundreds or thousands of frames ensuring their temporal continuity, consistency and dynamics. Our approach leverages a Markov chain framework with a recall mechanism, where each state represents a short-length VAE-GAN video generator. This setup enables the sequential connection of generated video sub-sequences, maintaining temporal dependencies and resulting in meaningful long video sequences.
\end{abstract}

  {\bf Keywords:}  Video generation, Generative Adversarial Networks, Variational Autoencoder, VAE-GAN, long-video generation, Markov chain.

\section{Introduction}
\label{sec-intro}

Videos are represented as sequences of image frames temporally connected by the motion occurring within the scene. Video generation aims to synthesize sequences of frames with high visual quality, ensuring temporal consistency and coherence while maintaining a realistic appearance. The core challenge is to model both spatial and temporal relationships in a way that is appealing, or at least acceptable, to the human vision system. Existing video generation methods attempt to leverage the relationships between frames, either by modeling the movement in the scene or by reducing the redundancy between consecutive frames \cite{aldausari2022video_review}.

Models relying on Generative Adversarial Networks (GANs), ranging from the Video-GAN (VGAN) to G$^3$AN \cite{clark2019dvdgan,saito2017tgan,vondrick2016vgan, tulyakov_mocogan_2018,  wang_g3an_2020}, typically decompose video information into separate processing streams to represent the variant and invariant aspects of the video, such as motion and content, similar to video classification models, as in \cite{BQN}. 
Models that incorporate inference mechanisms, such as AutoEncoders (AEs) or Variational Autoencoders (VAEs), have been shown to be more effective in constraining and controlling the generated video results, \cite{ Ardino2021c2m_ICCV,  babaeizadeh2021fitvid, Blattmann_2021_iPoke_ICCV, 2018VideoVAE_entropy, wang2018vid2vid, yan2021videogpt}. 

Many video generation models were developed starting from image generation algorithms 
\cite{Munoz_2021_tsgan, stylegan_v2022cvpr, sun_twostreamvan_2020,  tian2021mocoganhd}.
Hybrid VAE-GAN models have been proposed to combine the strengths of VAEs and GANs, mainly for image classification \cite{2015AAE, Larsen2016VAE/GAN, fei2020lvaegan} and also in the context of continual learning of images \cite{FeiBors_LGAA}. 
In this paper, we initially propose enabling a video GAN-based generator with an encoder mechanism, akin to a video VAE-GAN generator. This mechanism is employed separately in the content and movement spaces, where the movement is represented as successive frame differences. 
Our proposed method, called the Encoder GAN3 (EncGAN3), decomposes the video sequence into two streams: one representing content (temporally invariant parts) and the other representing movement (temporally variant parts). This separation allows for more effective modeling of both static and moving video characteristics, while the discriminator ensures the realism of the resulting video, as in the typical GAN-based generators. EncGAN3 consists of three  processing modules: Encoder, Generator and Discriminator, each trained independently to preserve the generation performance based on a random seed input. The Encoder decomposes the video sequence in the two streams representing content and movement, respectively. The Generator then integrates the content and movement streams at multiple scales to form the generated video stream, similarly to G$^3$AN \cite{wang_g3an_2020}. The Discriminator ensures the realism of both video content and motion.

Despite the rapid advancements in improving frame quality and video motion consistency in video generation, little progress has been made in generating longer video sequences. Simply extending existing models often leads to a rapid degradation in the quality of subsequent frames \cite{ge2022tatsLong}. Even with frame interpolation, high-quality results can only be achieved for up to 70-100 frames, equivalent to 2-4 seconds of video at standard frame rates (20-30 frames-per-second). Such limitations are particularly problematic for longer, more meaningful video sequences. However, videos that are meaningful to people and useful in various applications should be longer than that. 

Long-term video generation models \cite{SurveyLongVideo}, rely on training using short length videos while considering inference mechanisms for extending the generated temporal sequences with further frames by either using more sparse video representations \cite{stylegan_v2022cvpr, digan2022iclr}, or by training separate modules that capture temporally smooth latent codes within short sequences~\cite{ge2022tatsLong, tian2021mocoganhd,yang2022encgan}. While these approaches have achieved some success in generating random shape movements ({\em e.g.}, clouds or fire), they struggle to handle complex, structured motions like human activities, \cite{wang2023styleinv}.

In this paper, we propose a new approach for long video generation, initiated from short video sequences generated by EncGAN3. EncGAN3 models temporal relationships between individual frames, generating short sequences of 16 frames. By building upon these results we propose employing a recall mechanism that links the generated short video-clips while ensuring the movement continuity between consecutive clips. Using a Markov chain framework, consecutive video clips are treated as connected states. This approach, called Recall EncGAN3 (REncGAN3), enables the connection of short clips into extended sequences while requiring a fixed rather small memory for arbitrarily long videos.

This study brings the following contributions~:
\begin{enumerate}
\vspace*{-0.2cm}
\item A new hybrid VAE-GAN video generation method EncGAN3, which enables 
inference-based video generation.
\item Long video generation by using a recall mechanism through REncGAN3, by connecting short clips within a Markov chain, resulting in consistent and coherent sequences of hundreds and thousands of video frames.
\item Quantitative and qualitative results show the advantages of EncGAN3 and REncGAN3 with respect to the visual quality and diversity of the generated videos.
\end{enumerate}

The rest of the paper is organized as follows. The literature review of video generation  area is provided in Section~\ref{sec-litr}. The architecture and training of EncGAN3 are described in Section~\ref{sec-EncGAN3}. Further, the Markov chain representation and recall mechanism for long-term video generation are presented in Section~\ref{sec-recall}. Experiments and their discussion are provided in Section~\ref{sec-exps}, while the conclusions of this study are drawn in Section~\ref{sec-Con}.

\section{Related Works}
\label{sec-litr}

\subsection{Short-term Video Generation}

Early video generation methods focused on grouping together sequences of consecutive temporally changing images. 
Expanding 2D convolutional layers to the 3D to account for the temporal dimension was shown to be computationally expensive~\cite{eccv2018vidI3D}. Consequently, several architecture designs aimed at finding more efficient video data representations, \cite{Ardino2021c2m_ICCV,clark2019dvdgan, ho2022diffusion_iclr, Menapace2021_pvg_CVPR, saito2017tgan, stylegan_v2022cvpr,  sun_twostreamvan_2020, 
tulyakov_mocogan_2018,vondrick2016vgan,   wang_g3an_2020,    zhang2023tsvc_long}.

Video generation has drawn several useful insights from the field of video understanding. For example, multi-stream processing of video information, initially popularized in video understanding \cite{enc2017action, BQN}, was later adapted for video generation~\cite{vondrick2016vgan} and has since become a common format \cite{clark2019dvdgan}. 
Several models, such as the Temporal Generative Adversarial Nets (TGAN) \cite{saito2017tgan}, Motion and Content GAN (MoCoGAN) \cite{tulyakov_mocogan_2018},  VideoVAE \cite{2018VideoVAE_entropy}, and Temporal Shift GAN (TS-GAN) \cite{Munoz_2021_tsgan}, add a temporal processing module to the image decoder to create image sequences. 
Similarly, TwoStreamVAN~\cite{sun_twostreamvan_2020} uses an encoder to synthesize a motion stream that provides temporal information to a content stream, which then reconstructs video frames, \cite{yang2022encgan}. 
Most of these methods extend either VAE- or GAN-based image generation models to the video domain \cite{gupta2022rvgan, hong2023dagan, kumar2024posegan, kumar2023citygan, mittal2024designgan}. Among the methods that are not simple extensions of image generation models, G$^3$AN~\cite{wang_g3an_2020} employs a three-stream generator to reconstruct motion and content separately, then fuse them at multiple scales into the video reconstruction stream. More recently, diffusion models have also been successfully used for video generation, \cite{DiffVideo,ho2022diffusion_iclr, mei2023vidm128f_diffu}.

The video generation is computationally demanding because of having to estimate a large number of parameters on a 3D spatio-temporal grid. Some approaches reduce this burden by incorporating prior information to constrain the generated videos. From this perspective, video generation models can be classified by the type of required prior information, such as being conditioned on different types of semantic maps, by triggering the video generation from a given single frame, or by using several starting frames \cite{aldausari2022video_review}. The unconditional video generation is a special case of video synthesis that relies on the least prior information, by considering random noise as input, following the standard GAN framework \cite{clark2019dvdgan, saito2017tgan,  stylegan_v2022cvpr, tian2021mocoganhd, tulyakov_mocogan_2018, vondrick2016vgan, wang_g3an_2020, wang2023styleinv}. In contrast, conditional methods are often based on VAE architectures to benefit from its inference mechanism for better controlling of generated content \cite{2018VideoVAE_entropy, xu2023condi_vae}.  The development of conditional GAN-based video generation models~\cite{saito2020tganv2,spampinato2020rnn_cnn_condigan} was inspired by incorporating class information in GAN-based image generation~\cite{2018BigGAN_iclr}.
Conditional video generation constrains the output based on certain information, such as associating discrete labels~\cite{wang_imaginator_2020}, image features~\cite{sun_twostreamvan_2020,zhao2020motionTransfer_vaegan} or by means of interactive information \cite{Ardino2021c2m_ICCV, Blattmann_2021_iPoke_ICCV,Menapace2021_pvg_CVPR}.

\subsection{Existing long-term Video generation methods}

Most video generation methods model the dependency between consecutive frames aiming to ensure the temporal continuity together with frame quality. However, the computational complexity and memory limitations restrict the generated video length to typically 16 frames in many video generation models.
Recently, following the development of more efficient video representations, some video generation methods have been able to increase the frame resolution of the generated videos from 64 $\times$ 64 \cite{wang_g3an_2020, Menapace2021_pvg_CVPR, ho2022diffusion_iclr} to 128 $\times$ 128 \cite{Sky,saito2020tganv2, digan2022iclr, ge2022tatsLong}, 256 $\times$ 256 \cite{stylegan_v2022cvpr} and even to 1024 $\times$ 1024 in \cite{tian2021mocoganhd, stylegan_v2022cvpr}.  TGAN\_v2 \cite{saito2020tganv2} also combines the data produced by several sub-generators, aiming to achieve increased generated video resolution, instead of longer-temporal videos as it is proposed in Section~\ref{sec-recall}.

Despite improvements in generating high-resolution video frames, producing long sequences of temporally consistent and realistic frames remains a significant challenge \cite{li2024lvg_survey3}. Long-term video generation requires modeling temporal relationships across various scene regions, ensuring smooth continuity in the movement and other complex and yet realistic variations, such as changes in lighting or perspective, across many synthesized frames.
To capture such temporal dependencies, models need to be trained on videos with similar lengths to those being generated. However, the expensive computational and memory demands for longer training video lengths make this difficult. Forcing models trained on short videos to generate extended sequences via latent feature expansion or by simple frame interpolation results in rather poor results,
resulting in either quick deterioration of the frame quality or in artificial repeated movements, as discussed by \cite{ge2022tatsLong}.

Recent studies in long-term video generation have managed to extend the number of generated frames by increasing the generated video length from 64 frames \cite{tian2021mocoganhd} to 128 or 256 frames 
\cite{wang2023styleinv,digan2022iclr}, and even up to 1024 frames \cite{ge2022tatsLong} or longer \cite{nips2022dynamicLong,stylegan_v2022cvpr}, while in earlier generative models the number of generated frames for videos has been of around 10 to 32 frames \cite{Munoz_2021_tsgan,saito2017tgan, sun_twostreamvan_2020, tulyakov_mocogan_2018, vondrick2016vgan, wang_g3an_2020}.

Among the long-term video generation models, StyleGAN-V~\cite{stylegan_v2022cvpr} uses a more computationally efficient video representation. However, during training, StyleGAN-V tends to reuse previous motion information, eventually leading to repetitive movement when aiming to generate complex movements, as in human activities~\cite{stylegan_v2022cvpr}. 
Similarly, DIGAN \cite{digan2022iclr} incorporates spatio-temporal modeling to produce more coherent human action videos of up to 128 frames without significant quality loss. 
Meanwhile, TATS \cite{ge2022tatsLong} captures latent codes for temporally related frames, generating videos of up to 102 frames. While effective at creating longer videos, TATS often results in repetitive or unrealistic human action movements after a certain number of generated frames.
However, these methods do not fully learn the long-term temporal dependencies needed for generating seamlessly extended videos. Instead, they primarily aim to ensure smooth transitions between consecutive frames, with quality degradations  progressively increasing with the video length. Moreover, they struggle to generate complex and coordinated human activities.

In this study, we leverage VAE's inference ability in video streams by proposing a Recall mechanism, described in Section~\ref{sec-recall}, to enhance the EncGAN3 generator. This enables flexible constraint adjustment by connecting video clips, ensuring the video continuity based on the information from previous clips, ensuring the generation of long-term videos by means of generating meaningfully connected clips. 

\section{The Encoding GAN3 (EncGAN3)}
\label{sec-EncGAN3}


A video sequence  ${\cal V}$ is considered as a series of image frames 
${\cal V} = \{ {\bf x}_i \mid i=1,\ldots, T \}$ which show the dynamic scene of a certain location, occurring within a certain time. The generated video sequence  $\widehat{\cal V}$  is also defined by a sequence of synthesized frames $\widehat{\cal V} = \{ \widehat{\bf x}_i \mid i=1,\ldots, T \}$. In this section, we consider short video clips with $T < 100$ for training as well as also for generation, while in Section~\ref{sec-recall} we discuss an approach for generating longer videos. By considering the frames in a video sequence as being related to each other, assuming that they correspond to the same dynamic scene, the generation of a video sequence can be defined as a Markov chain, where each generated frame depends upon all the previously generated ones, in a temporal sequence~:
\begin{equation}
p (\widehat{\cal V}) = \prod_{i=1}^T p(\widehat{\bf x}_i) =
\prod_{i=1}^T p(\widehat{\bf x}_i \mid \widehat{\bf x}_0, \widehat{\bf x}_1, \ldots, \widehat{\bf x}_{i-1} ),
\label{MarkovEncG}
\end{equation}
where $p (\widehat{\cal V})$ represents the probabilistic representation of the generated video sequence and $p(\widehat{\bf x}_i)$ describes the probabilistic representation of $\widehat{\bf x}_i$. In the following, we consider the dependency between consecutive frames as defined by a matrix~:
\begin{equation}
\widehat{\bf v}_i = | \widehat{\bf x}_i \ominus \widehat{\bf x}_{i-1} |,
\label{FrameDiff}
\end{equation}
where $\ominus$ represents the pixel-wise differences between two image frames and is extended to all synthesized frames $i=2,\ldots,T$. $\widehat{\bf v}_i$ represents the movement estimation in the generated video. This leads to approximate
Eq.~\eqref{MarkovEncG} through~:
\begin{equation}
p (\widehat{\cal V}) = 
\prod_{i=2}^T p(\widehat{\bf x}_i \mid \widehat{\bf v}_i, \widehat{\bf x}_{i-1} ),
\label{MarkovGen}
\end{equation}
where the generation of the video frames is following a recursion starting from an initially generated frame $\widehat{\bf x}_1$, which is considered as representing the scene content, together with the movement generative representation 
$\{\widehat{\bf v}_i \mid i=2,\ldots, T \} $.

\subsection{Architecture}
\label{sec-ArchEncGAN3}

\begin{figure*}[h]
    \centering
    \includegraphics[width=1\linewidth]{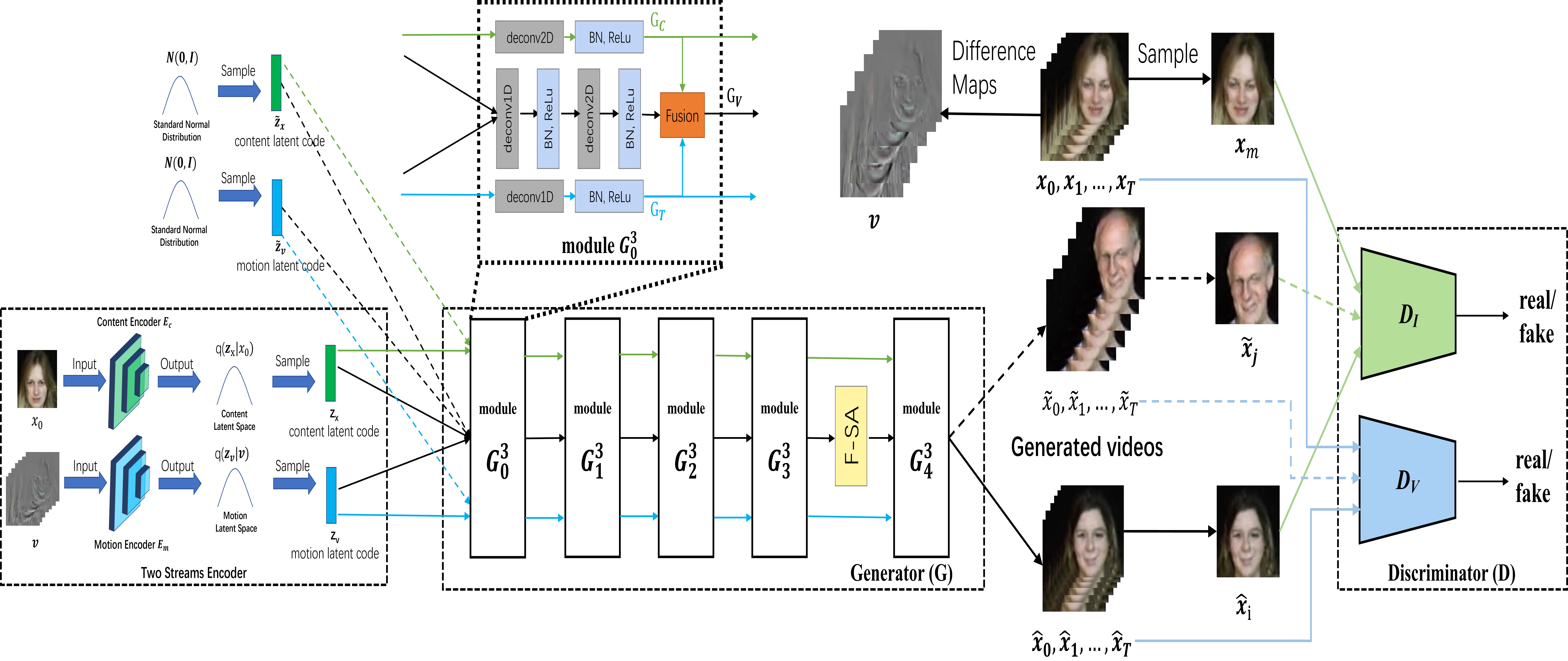}
    \caption{The architecture of EncGAN3: two Encoders ($Enc$), three-stream Generator ($G$) and two Discriminators ($D$) for content and movement.}
    \label{fig:EncGAN3}
\end{figure*}

Despite the generation abilities of GANs, their data synthesis cannot be controlled effectively and sometimes results in data with unexpected artifacts.
To address such issues we propose to enable a GAN-based video generator with an encoder in a model called the Encoding GAN3 (EncGAN3). The architecture of EncGAN3, shown in Figure~\ref{fig:EncGAN3}, consists of three main modules~: Encoder, Generator and Discriminator. During the training, the video is decomposed into content and movement, where two encoders are used for representing their corresponding latent spaces, similar to the TwoStreamVAN \cite{sun_twostreamvan_2020}. 
The latent space features corresponding to the difference maps from the video sequence are compressed to produce the motion latent code while in parallel the encoder extracts the features corresponding to the initial frame, as the content representation. Then, the Generator uses the latent space for producing the content frame and the corresponding movement stream and then both streams are fused forming the generated video. 
The two mechanisms, of inference and generation, are matched in their latent spaces corresponding to the scene content and movement. 

The generator's architecture for processing the three streams consists of five stacked modules denoted as $\{ G^3_i \mid i=0,\ldots,4\}$ and a factorized self-attention (F-SA) module, implemented before the $G^3_4$ module, as in G$^3$AN \cite{wang_g3an_2020}, as shown in Figure~\ref{fig:EncGAN3}.
During the training, the Generator is fed with both latent codes from the two encoders as well as with noise sampled from the standard normal distribution. 
The video frames are reconstructed by recursively adding the generated movement $\widehat{\bf v}_i$, frame by frame, to the corresponding generated frames. This process is implemented at different scales, after each module $G^3_i$, $i=0,\ldots,4$ starting with the frame generated from the content latent variables. 
The F-SA module implements temporal-wise self-attention (SA), followed by a spatial-wise SA which benefits the generator $G$ to identify and model distinct areas by using cues from the spatial-temporal features. This module ensures the consistency of the generated video originating from the latent space codes while considering the variability enabled by using the random seed of the Generator. 
The data flow processing pipeline ends with a two-stream Discriminator which judges the realism of generated videos, by comparing their probabilistic representations to those of randomly selected frames from real videos.
After the training, EncGAN3 can generate videos by using concurrently the Encoder and Generator, through a hybrid VAE-GAN generative process. The model can also be adjusted, after decoupling the Encoder, and only using the Generator as in GAN generation.

\subsection{Training }
\label{sec-loss_enc}       

The EncGAN3 is a hybrid model employing characteristics from both VAEs and GANs, as described in the previous section. Each of the component modules, Encoder, Generator and Discriminator, is trained separately and has its own loss function. The training of EncGAN3 is explained in the following. 

The loss function of the two-stream Encoder, denoted by $L_{Enc}$, is given by~: 
\begin{align}
    L_{Enc}  = &  \sum_{i=1}^N \left(  \| {\bf x}_{i0} - \widehat{\bf x}_{i0} \|^2  \right. \notag \\
         & \left. + \sum_{j=1}^{T_i} \| {\bf x}_{ij} - \widehat{\bf x}_{ij} (\widehat{\bf x}_{i,j-1}, \widehat{\bf v}_{i,j-1}) \|^2 \right)  \notag \\
        & + D_{KL} (q_{{\theta}_{\bf x}} ({\bf z}_{\bf x} \,|\, {\bf x}) \| p ({\bf z}_{\bf x}))  \notag \\
        & + D_{KL} (q_{{\theta}_{\bf v}} ({\bf z}_{\bf v} \,|\, {\bf v}) \| p ({\bf z}_{\bf v})),
    \label{Obj_Enc}
\end{align}
where we consider $N$ video clips of $T_i$, $i=1,\ldots,N$ frames each, for training. The difference maps $\widehat{\bf v}_{ij}$ are the reconstructions of the movement in the predicted videos, representing differences between consecutive predicted image frames. The image reconstructions $\{ \widehat{\bf x}_{ij} \}_{j=0}^{T_i}$ are calculated recursively using the frame differences $\{ \widehat{\bf v}_{ij} \}_{j=1}^{T_i}$:
\begin{equation}
    \begin{aligned}
        \widehat{\bf x}_{ij}  =  \widehat{\bf x}_{i,j-1} \oplus \widehat{\bf v}_{i,j-1} ,\; j=1,\ldots,T_i,
    \end{aligned}
    \label{VideoReconstr}
\end{equation}
where $\oplus$ represents pixel-wise image addition, for $i=1,\ldots,N$. The reconstructions $\{ \widehat{\bf x}_{ij} \}_{j=0}^{T_i}$ are made as close as possible to their corresponding original frames $\{ {\bf x}_{ij} \}_{j=0}^{T_i}$ by the content and motion encoders.  
While the content encoder models ${\bf x}_{i0}$ by $\widehat{\bf x}_{i0}$, the motion encoder focuses on reconstructing the frame differences $\{ \widehat{\bf v}_{ij}\}_{j=1}^{T_i}$, for representing the motion. The joint optimization of content and motion encoders achieves better reconstruction results than optimizing them separately.
By minimizing $L_{Enc}$, the two encoders implementing the variational distributions $q_{{\theta}_{\bf x}} ({\bf z}_{\bf x} \,|\, {\bf x})$ and $q_{{\theta}_{\bf v}} ({\bf z}_{\bf v}  \,|\, {\bf v})$, create feature representations of the content ${\bf z}_{\bf x}$, and motion ${\bf z}_{\bf v}$.
The KL divergence terms $D_{KL}$ ensure that the probabilities of the latent variables associated with the content ${\bf z}_{\bf x}$ and motion ${\bf z}_{\bf v}$, generated by the two encoders of parameters ${\theta}_{\bf x}$ and ${\theta}_{\bf v}$, are consistent with the Normal distributions, $p ({\bf z}_{\bf x})$ and $p ({\bf z}_{\bf v})$.

 The loss function of the Generator $L_G$ combines the VAE and GAN losses:
\begin{align}
         & L_G  =  \mathbb{E}_{\widehat{\bf x}_n \sim G({\bf z}_{\bf x},{\bf z}_{\bf v})} 
          \log [D(\widehat{\bf x}_n)] \notag \\
         &  \hspace*{0.8cm}  + \mathbb{E}_{\tilde{\bf x}_n \sim G(\tilde{\bf z}_{\bf x},\tilde{\bf z}_{\bf v})} \log [D(\tilde{\bf x}_n)] \notag \\
         &  \hspace*{0.8cm}  +  \mathbb{E}_{{\bf z}_{\bf x}\sim p({\bf z}_{\bf x}), {\bf z}_{\bf v} \sim p({\bf z}_{\bf v})} \log [D(G({\bf z}_{\bf x},{\bf z}_{\bf v}))]  \label{Obj_Gen} \\ 
         & \hspace*{0.8cm}  + \mathbb{E}_{\tilde{\bf z}_{\bf x} \sim {\cal N}(0,{\bf I}), \tilde{\bf z}_{\bf v} \sim {\cal N}(0,{\bf I})} \log  [D(G(\tilde{\bf z}_{\bf x},\tilde{\bf z}_{\bf v}))]  \notag \\
         & + \sum_{i=1}^N  \left( \| {\bf x}_{i0} - \widehat{\bf x}_{i0} \|^2 + \sum_{j=1}^{T_i} \| {\bf x}_{ij} - \widehat{\bf x}_{ij} (\widehat{\bf x}_{i,j-1}, \widehat{\bf v}_{i,j-1}) \|^2 \right) \notag 
\end{align}
where the terms from the last line from the Right Hand Side (RHS) of Eq.~\eqref{Obj_Gen} are the same to those from the first row of the RHS from Eq.~(\ref{Obj_Enc}). 
The data seeds in Eq.~\eqref{Obj_Gen} are the latent variables of the content ${\bf z}_{\bf x}\sim p({\bf z}_{\bf x})$ and that of the movement $ {\bf z}_{\bf v} \sim p({\bf z}_{\bf v}) $, modeled by the corresponding encoders, as well as those produced by the random generators of the GAN network for content, $ \tilde{\bf z}_{\bf x} \sim {\cal N}(0,{\bf I})$, and movement, $\tilde{\bf z}_{\bf v} \sim {\cal N}(0,{\bf I})$. The content $\widehat{\bf x}_0$ and movement $\widehat{\bf v}$ are reconstructed considering the latent codes created by the encoders, ${\bf z}_{\bf x}$ and ${\bf z}_{\bf v}$, as well as those generated by the random generators $ \tilde{\bf z}_{\bf x}$ and  $\tilde{\bf z}_{\bf v}$. 
The second and fourth components from the RHS  of Eq.~\eqref{Obj_Gen}  use the GAN generator, where the codes $\tilde{\bf z}_{\bf x}$ and $\tilde{\bf z}_{\bf v}$ are sampled from the normal distributions for content and motion, respectively.

The loss functions of the two-stream Discriminator, $L_{D_I}$ and $L_{D_V}$, correspond to the adversarial losses, similar to those used in MoCoGAN \cite{tulyakov_mocogan_2018}, G$^3$AN \cite{wang_g3an_2020} and TwoStreamVAN \cite{sun_twostreamvan_2020}. The two streams, corresponding to the content and movement information, each with its own Discriminator, are trained in parallel. The loss function of the image-stream Discriminator $L_{D_I}$ is given by~:
\begin{equation}
    \begin{aligned}
         L_{D_I} = & \hspace*{0.1cm}   \mathbb{E}_{{\bf x}_n \sim p ({\bf x})} \log [D({\bf x}_n)]  \\
         &  + \mathbb{E}_{\widehat{\bf x}_n \sim G({\bf z}_{\bf x},{\bf z}_{\bf v})} \log [1-D(\widehat{\bf x}_n)]   \\
        &+ \mathbb{E}_{\tilde{\bf x}_n \sim G(\tilde{\bf z}_{\bf x},\tilde{\bf z}_{\bf v})} \log [1-D(\tilde{\bf x}_n)] , \\
    \end{aligned}
    \label{LossLDI}
\end{equation}
where ${\bf x}_n \sim p ({\bf x})$ is the real image content, $\widehat{\bf x}_n$ is generated from the latent codes representing the image content, and $\tilde{\bf x}_n$ is generated using the standard Gaussian distribution. Images $\widehat{\bf x}_n$ and $\tilde{\bf x}_n$ are randomly sampled from the video clips, where $n \in \{0,\ldots,T_i\}$.

The video-stream Discriminator $L_{D_V}$ loss function is defined as~:
\begin{align}
     L_{D_V} = & \hspace*{0.1cm} \mathbb{E}_{{\bf x}_{0:T} \sim p ({\bf x}_{0:T})} \log [D({\bf x}_{0:T})] \label{LossLDV} \\
     &  + \mathbb{E}_{\widehat{\bf x}_{0:T} \sim p (\widehat{\bf x}_{0:T})} \log [1-D(\widehat{\bf x}_{0:T})] \notag \\
     & + \mathbb{E}_{\tilde{\bf z}_{\bf x} \sim 
     {\cal N} (0,{\bf I}), \tilde{\bf z}_{\bf v} \sim 
     {\cal N} (0,{\bf I})} \log [1-D(G(\tilde{\bf z}_{\bf x},\tilde{\bf z}_{\bf v})] , \notag
\end{align}
where ${\bf x}_{0:T} = \{ {\bf x}_j \}_{j=0}^{T-1}$ and $\widehat{\bf x}_{0:T} = \{ \widehat{\bf x}_j \}_{j=0}^{T-1}$ represent the real videos and their reconstructions, while $p ({\bf x}_{0:T})$ and $ p (\widehat{\bf x}_{0:T})$ are their probabilities. The second term from RHS of Eq.~\eqref{LossLDV} represents the evaluations of the discriminator on the videos generated from the latent codes of the Encoder, while the third term represents the evaluations on videos generated by the random seed generator.
The reconstruction of the video frames  $\widehat{\bf x}_{0:T}$ depends on firstly reconstructing the content $\widehat{\bf x}_0$ and then the frame differences $\{ \widehat{\bf v}_j \}_{j=1}^{T-1}$, as in Eq.~\eqref{FrameDiff}, where $\widehat{\bf x}_0$ represents $\widehat{\bf x}_{i0}$ and $\widehat{\bf v}_j$ is used instead of $\widehat{\bf v}_{ij}$ for simplification.

During the training, first the Discriminator is updated by optimizing $L_{D_I}$ and $L_{D_V}$ using Eq.~(\ref{LossLDI}) and Eq.~(\ref{LossLDV}), then the Encoder using $L_{Enc}$ from (\ref{Obj_Enc}), and eventually the Generator $L_G$, according to 
Eq.~(\ref{Obj_Gen}).

\section{Long-term Video Generation}
\label{sec-recall}

The generated video length by the EncGAN3 is restricted to less than 100 frames, even after expanding the temporal latent vector size. This represents a serious limitation in representing long-term video information. To overcome this limitation, in this section we introduce a new framework, called the recall mechanism. 
The length of the generated video depends on the length of videos used in the training, which is also constrained by the available computational memory. The recall mechanism considers the interdependence between consecutive video clips, allowing the modeling of longer videos displaying continuous and consistent movement while only using limited computational and memory resources.

\subsection{General framework for long-temporal video generation}
\label{REncGAN3}


Most video generation methods developed so far produce short-term video clips, usually showing just a short instance of movement which does not cover more than one second of video time. However, representative videos useful for practical applications should be much longer. 
The generation of long-duration videos showing realistic movement is a very challenging task due to having to keep the continuity and consistency of video movement for significant lengths of time. 

Although the generated video length can be significantly increased through interpolation, it does not result in any new meaningful variations or increases in the diversity of the generated video and usually ends up with a poor visual outcome.
Limitations in videos showing movements of realistic body actions are evident in the results produced by the few long-term video generation algorithms proposed so far such as those by \cite{stylegan_v2022cvpr,digan2022iclr,ge2022tatsLong}.
For example, the generated longer video sequences display repetitive movements in the results by some models \cite{ge2022tatsLong,stylegan_v2022cvpr},   which actually do not correspond to realistic videos.

Similarly to other video generation methods, EncGAN3  produces short video sequences, and in the experimental results we consider the generation of video sequences of $T=16$ frames by EncGAN3, representing less than 1 second of HD video. 
Nevertheless, EncGAN3 has a major advantage over G$^3$AN \cite{wang_g3an_2020} or other GAN-based video generators, by being empowered with an inference mechanism. 
In the following, we expand the EncGAN3 model, defined in Section~\ref{sec-EncGAN3}, for generating probabilistic dependencies enabled by Markov chains, by enforcing the continuity between successive video clips. 
A long video $\widehat{\bf y}_{1:T}$ is created by recursively connecting pairs of shorter video clips $\widehat{\bf x}_{j,1:T_c}$, $j=1,\ldots,N$ for the entire sequence of $N$ generated video clips, where $T \gg T_c$. The continuity between consecutive segments of a long video is ensured by considering a reference frame $r_j$ from one video clip $j$, which is used for enabling the continuity and smoothness transition to the next video segment $j+1$, where $j=1,\ldots,N-1$. This mechanism continuously links successive generated video clips leading to a long video sequence showing coherence in movement.
The generation of long video sequences is represented as a Markov chain, where we first simplify the dependency of entire video clips to that of a reference frame~:
\begin{align}
p (\widehat{\bf y})  & = \prod_{j=1}^N p(\widehat{\bf x}_{j,1:T_c}) \notag \\
& = p(\widehat{\bf x}_{1,1:T_c}) 
\prod_{j=2}^N p(\widehat{\bf x}_{j,1:T_c} \mid \widehat{\bf x}_{j-1,1:T_c} ) \notag \\
  & \approx p(\widehat{\bf x}_{1,1:T_c})
\prod_{j=2}^N p(\widehat{\bf x}_{j,1:T_c} \mid \widehat{\bf x}_{j-1,r} ) \; ,
\label{LongGen}
\end{align}
where we consider the notation $\widehat{\bf x}_{j,1:T_c} $ 
for the short video sequences of length $T_c$ for $j=2,\ldots,N$, generated according to the description from Section~\ref{sec-EncGAN3}, while $\widehat{\bf x}_{j-1,r}$, represents the reference frame from $j-1$th video clip, where we consider the same reference index $r \in \{1, \ldots, T_c\}$ for all video clips, when generating the long video sequence $\widehat{\bf y}$, where we also drop the video clip index from the reference frame. Individual frames from each video clip eventually become part of the longer video sequence $\widehat{\bf y}$. 

 The dependency of a certain video segment on all previous frames through a Markov chain is reduced to the dependency on only the previous video clip, which is part of a long-temporal video stream. Furthermore, this is reduced to the dependency of a video clip $j$ on a single reference 
 frame $\widehat{\bf x}_{j-1,r}$ from the previous video clip, thus enabling a connection link between successions of video clips, eventually ensuring the coherence in the generated long video sequence. By generating successive video clips using EncGAN3 and then considering certain reference frames for linking them one to another we implicitly assume that we have $T_c-r$ overlapping frames between successive video segments, where $T_c$ represents the number of frames in a video clip. 

The proposed long-term video generation framework relies on initially generating short video clips, $j=1,\ldots,N_c$, as described in Section~\ref{sec-EncGAN3}. Unlike in EncGAN3, where we consider the first frame $\widehat{\bf x}_{j,1}$ as the content information, in the long-term video generation we consider a reference frame $\widehat{\bf x}_{j,r}$, located within the video clip $0 < r < T_c$, while we also use the movement representations 
$\widehat{\bf v}_{j,i}$ for $j=1,\ldots,N$, and $i=1,\ldots, T_c-1$, as in EncGAN3.
The generation of the video clips is similar to that of EncGAN3, but considering the reference $\widehat{\bf x}_{j,r}$ as the content frame together with the generated movement for calculating the other frames from clip $j$~:
\begin{align}
   \widehat{\bf x}_{j,i} & =  \widehat{\bf x}_{j,i+1} \ominus \widehat{\bf v}_{j,i+1} ,\; i=r-1,\ldots,1 ,
       \label{VideoReconstr_subs1} \\
   \widehat{\bf x}_{j,i} & =  \widehat{\bf x}_{j,i-1} \oplus \widehat{\bf v}_{j,i} , \; i=r+1,\ldots,T_c .
        \label{VideoReconstr_subs2}
\end{align}
Equations~\eqref{VideoReconstr_subs1} and \eqref{VideoReconstr_subs2} are used to generate the initial and final parts of a video clip, where `$\ominus$' and `$\oplus$' represent the subtraction and addition, respectively, of frame differences representing movement. These frames are generated recursively for $j=1,\ldots,N$ video clips. 
By replacing Eq.~\eqref{VideoReconstr} from EncGAN3 with Eq.~\eqref{VideoReconstr_subs1} and \eqref{VideoReconstr_subs2} for the clip generation,
we build the Markov chain, connecting each video clip to the next one. 
However, for building a long sequence, we should ensure that each generated video clip is smoothly connected to both its previous and next video clips by enforcing the continuity across the entire sequence, benefiting the modeling of temporal consistency in the long-term generated video.
Note that when training with videos longer than 16 frames (e.g., extending to 20 frames), there is a dramatic increase in GPU memory requirements (as shown in Tab. \ref{tab:memory_cost}), which directly limits the training video length. To overcome this, our proposed recall mechanism decomposes a long video into multiple temporally coherent sub-sequences, each requiring only a fixed amount of GPU memory. By connecting these sub-sequences using a Markov chain dependency on a designated reference frame, the overall memory usage remains constant regardless of the total video length, thereby enabling the generation of hundreds or even thousands of frames. Although this approach does not reduce the algorithmic time complexity or modify the storage, it effectively circumvents in practice the GPU memory bottleneck.
By enabling the temporal consistency in consecutive frames, we drop the frames generated in the second part by Eq.~\eqref{VideoReconstr_subs2} for a video clip $j$, and replace them with the frames generated by Eq.~\eqref{VideoReconstr_subs1} from the next video clip $j+1$. 
The exception is when adding the final video clip when these frames are considered as the ending part of the longer generated video $\widehat{\bf y}$.
Eventually the long video is generated by concatenating $r$ frames from each video clip, generated according to Eq.~\eqref{VideoReconstr_subs1}, with the last $T_c - r$ frames being generated according to Eq.~\eqref{VideoReconstr_subs2}, as~:
\begin{align}
 \widehat{\bf y}_{i + (j-1) * r} & = \widehat{\bf x}_{j, i}, \; i=1,\ldots,r, \; j=1,\ldots,N  
   \label{VideoReconstr_longlen} , \\
\widehat{\bf y}_{i + N * r} & = \widehat{\bf x}_{N_c,i-1} \oplus \widehat{\bf v}_{N,i}  \; i=r+1,\ldots,T_c \; .
\end{align}
The resulting video has a length of $(N-1)*r+ T_c$ frames.
The size of the long-term generated video can be different from that of the training video set, which is assumed to be long in order to ensure the richness of the training data for the long-term video generator.
In the following, we describe the long-temporal video training algorithm and the procedure for choosing the reference frame.

\subsection{The recall mechanism}
\label{TrainRecall}

\begin{figure*}[h]
    \centering
    \includegraphics[width=1\linewidth]{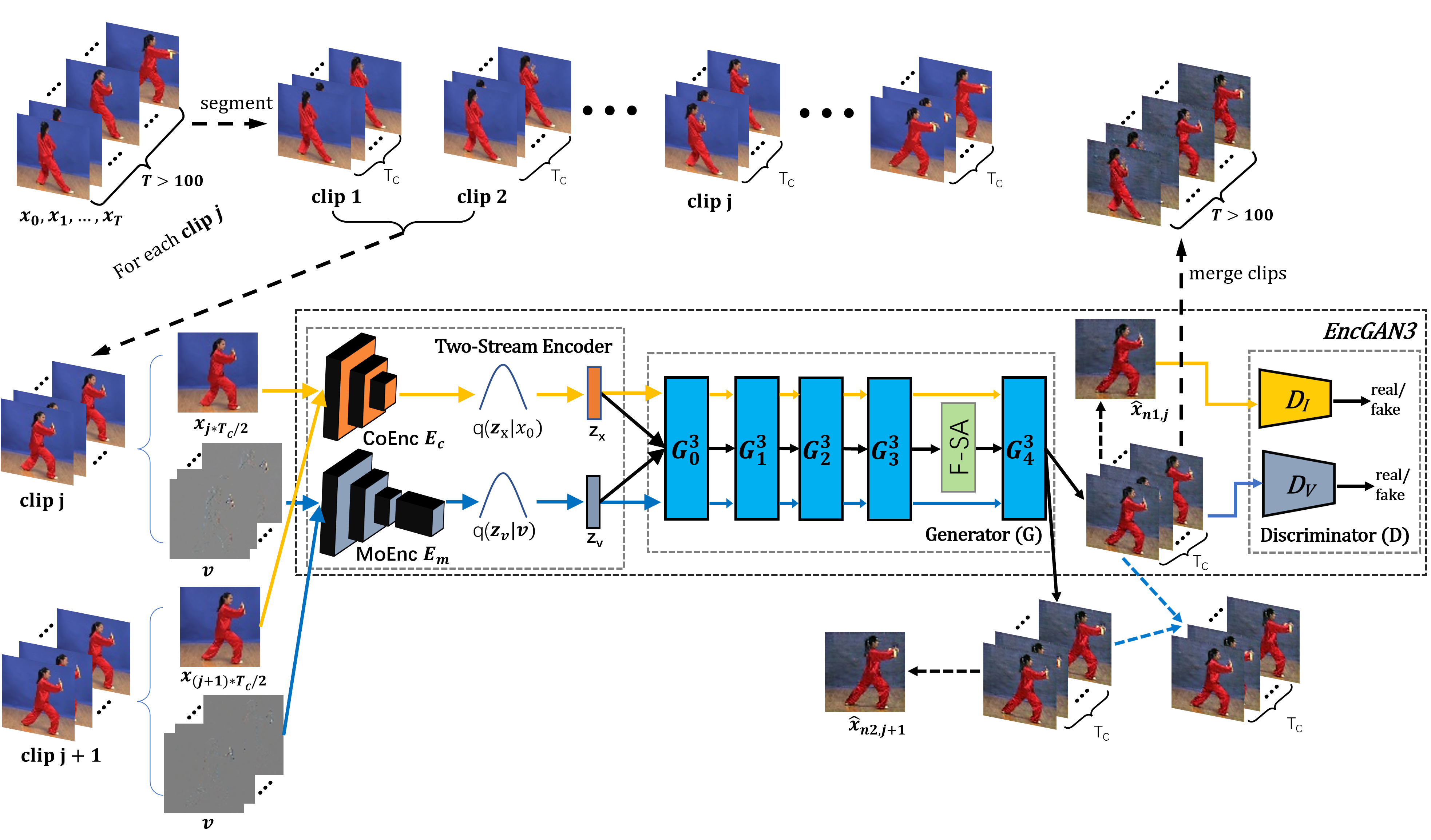}
    \caption{Illustration for the Recall EncGAN3 (REncGAN3) long-term video generation method.}
    \label{fig:REncGAN3}
\end{figure*}

In this section, we describe the Recall EncGAN3 (REncGAN3), which employs the recall mechanism to the EncGAN3, described in Section~\ref{sec-EncGAN3}, for enabling the generation of long-temporal video sequences. 
During the training, we jointly optimize the Encoder and Generator in REncGAN3 instead of training them separately as in EncGAN3 for a better performance in long video generation. 
In the recall mechanism, we make two changes on the way how the data is processed in the two-stream Encoder and then on how the video data is processed by the Video Discriminator. 
For the former change, we segment the long video sequences into overlapping video clips while for the latter one, we consider the reconstruction of the individually generated clips together with their merged video streams for the consistency checking by the Video Discriminator. This Video Discriminator mechanism ensures that the training enforces the video continuity and consistency between the consecutive segments of the long-temporal video sequence, \cite{yang2023R1enc}.

The long-term video generation REncGAN3 training is illustrated in Fig.~\ref{fig:REncGAN3}. The video sequences used for training are considered as being at least hundreds of frames long. These training long sequences are split into sets of consecutive $T_c$ frame clips. The clips are considered both individually and in pairs of two consecutive video clips. The consecutive video clips are all considered to be of the same size while having a certain number of overlapping frames, which are considered in order to model the continuity and consistency of the long-term video~:
\begin{align}
  {\bf y} = \bigcup  \{ & {\bf x}_{1:T_c} , {\bf x}_{r:r+T_c}, {\bf x}_{2r:2r+T_c} , \ldots, \notag \\
  & {\bf x}_{(N-1)r:(N-1)r+T_c}  \}, 
  \label{VideoMakeUp}
\end{align}
where  ${\bf y}$ represents a training long video sequence, $N$ is the number of video segments, and $r$ represents the reference defining the number of overlapping frames in the training sequence, assumed to be  $T_c-r$, identical in all video segments. 
According to the empirical experiments the best results are provided when the reference frame is $r=T_c/2$, resulting in $T_c/2$ overlapping frames between consecutive clips.

REncGAN3 is trained to utilize the connection information from the video clips used in the training for generating connected clips. 
Initially, as in the EncGAN3, the video clips generated are of $T_c$ frames in length.
These clips are then merged to become long videos by enabling continuous and coherent frame information by means of the reference frames $\{ {\bf x}_{i,T_c+r}\}_{i= 0}^{N-1}$, leading to overlaps between consecutive clips.
Each two generated clips are merged to create another, longer clip, and all such groups of three video sequences are considered as the input to the Video Discriminator $D_V$ while the two randomly sampled frames $\widehat{\bf x}_{n1,j}$ and $\widehat{\bf x}_{n2,j+1}$ are the inputs of the Image Discriminator $D_I$, as illustrated in  Fig.~\ref{fig:REncGAN3}.
Then, adversarial learning, characterizing the training of GAN models, enforces the generator to enable REncGAN3 to create connected clips.  
Actually, REncGAN3 requires only minor structural modifications from EncGAN3, resulting in only a small additional memory requirement for enabling long-term video generation. And the extra memory requirement remains unchanged for generating long videos of arbitrary length. The training of REncGAN3 is different from that of EncGAN3, by jointly training the Encoder and Generator, among other changes. The training for REncGAN3 also requires changes of the loss functions of the Encoder, Generator and Discriminator. The following loss function $L_{REncG}$ is used for training REncGAN3~: 
\begin{equation} 
    \begin{aligned}
    L_{REncG} & = \sum_{m=1}^{N_L} \sum_{j=1}^{N_C} \| {\bf x}_{m,Tc/2,j} - \widehat{\bf x}_{m,Tc/2,j} \|^2 \\
   &  + \sum_{m=1}^{N_L} \sum_{j=1}^{N_C}   \sum_{i=1}^{T_c-1} \| {\bf x}_{m,i,j} - \\
   & \hspace*{2.5cm} \widehat{\bf x}_{m,i,j} (\widehat{\bf v}_{m,i,j},\widehat{\bf x}_{m,T_c/2,j}) \|^2    \\
        & +  D_{KL} (q_{{\theta}_{\bf x}} ({\bf z}_{\bf x}\mid{\bf x}) \| p ({\bf z}_{\bf x}) ) \\
       & + D_{KL} (q_{{\theta}_{\bf v}} ({\bf z}_{\bf v}\mid{\bf v}) \| p ({\bf z}_{\bf v}) ) \\
      &   - \mathbb{E}_{{\bf z}_{\bf x}\sim q_{{\theta}_{\bf x}} ({\bf z}_{\bf x}\mid{\bf x}), {\bf z}_{\bf v} \sim q_{{\theta}_{\bf v}} ({\bf z}_{\bf v}\mid{\bf v})} \log [D(G({\bf z}_{\bf x},{\bf z}_{\bf v}))] \\
        & - \mathbb{E}_{\widehat{\bf x}_n \sim G({\bf z}_{\bf x},{\bf z}_{\bf v})} \log [D(\widehat{\bf x}_n)] ,
    \end{aligned}
    \label{LossREncG}
\end{equation}
where we consider $N_L$ long-term videos, each split into $N_C$ overlapping clips, each containing $T_c$ frames, ${\bf x}_{m,i,j} $ represents an image frame while $\widehat{\bf x}_{m,i,j} $ is its reconstruction, $\widehat{\bf v}_{m,i,j}$ the reconstruction of the movement, as the difference between consecutive frames, associated with the frame $i$ from clip $j$ from the long-term video $m$. Meanwhile, $\{{\bf z}_{\bf x}, {\bf z}_{\bf v}\}$ represent the latent spaces of the content and movement, modeled by the encoders $E_c$ and $E_m$, respectively. 
The reconstructions $\{ \widehat{\bf x}_{ij} \}_{i=1}^{T_c}$ are made as close as possible to their corresponding original frames $\{ {\bf x}_{ij} \}_{j=1}^{T_c}$ by the content and video reconstruction errors corresponding to the first and second terms in Eq.~\eqref{LossREncG}.
The two KL divergence terms $D_{KL}$ ensure that the probabilities of the latent variable associated with the content ${\bf z}_{\bf x}$ and motion ${\bf z}_{\bf v}$, generated by the two encoders of parameters ${\theta}_{\bf x}$ and ${\theta}_{\bf v}$, are normal distributions, namely $p ({\bf z}_{\bf x})$ and $p ({\bf z}_{\bf v})$.
By minimizing $L_{REncG}$, the two encoders implementing the variational distributions $q_{{\theta}_{\bf x}} ({\bf z}_{\bf x} \,|\, {\bf x})$ and $q_{{\theta}_{\bf v}} ({\bf z}_{\bf v} \,|\, {\bf v})$, encode the feature representations of content ${\bf z}_{\bf x}$ and motion ${\bf z}_{\bf v}$, for enabling a better data reconstruction by the Generator.  

Initially, we have the EncGAN3 discriminators from Eq. \eqref{LossLDI} and \eqref{LossLDV} for producing short-term video clips.
For the two-stream Discriminator of the long-term video generator REncGAN3, each stream is trained and both Discriminators are optimized in parallel, similar to the EncGAN3 but dropping the random number generators and also considering the merging of consecutive video clips in the training, which leads to making up the long-term video.
For the two-stream Discriminator, after considering the joint training of the Encoder and Generator through $L_{REncG}$ from Eq.~\eqref{LossREncG}, unlike in the Discriminator training for EncGAN3 from Equations~\eqref{LossLDI} and \eqref{LossLDV}, we drop the random number generators and obtain the following loss functions~:
\begin{align}
    L_{D_I,R}  = & \sum_{i=1}^N \left\{ - \mathbb{E}_{{\bf x}_i \sim p ({\bf x}_{T_c})} \log [D({\bf x}_i)] \right. \notag \\
        & \left. - \mathbb{E}_{\widehat{\bf x}_i \sim G({\bf z}_{\bf x},{\bf z}_{\bf v})} \log [1-D(\widehat{\bf x}_i)] \right\},
    \label{LossLDI2} \\
    L_{D_V, R1}  = & - \mathbb{E}_{{\bf x}_{0:T} \sim p ({\bf x}_{0:T})} \log [D({\bf x}_{0:T})] \notag \\
        & - \mathbb{E}_{\widehat{\bf x}_{0:T} \sim p (\widehat{\bf x}_{0:T})} \log [1-D(\widehat{\bf x}_{0:T})],
    \label{LossLDV2}
\end{align}
where $L_{D_I,R}$ and $L_{D_V,R1}$ are the loss functions for training the content and the video discriminators after dropping the random number  generator components.  ${\bf x}_i$ is a frame sampled from the real video clip, whose probability is considered to be $p({\bf x}_{T_c})$ and $\widehat{\bf x}_i$ is from the video generated by latent codes, assuming that the video sequence is split into $N= \lfloor T/T_c \rfloor$ video clips, while ${\bf x}_{0:T}$ and $\widehat{\bf x}_{0:T}$ represent the original and the generated long-term video sequences of $T$ frames. When considering the merging of consecutive video clips, we replace $L_{D_V, R1}$ from Eq.~\eqref{LossLDV2} with the following loss function~: 
\begin{equation}
\begin{aligned}
     L_{D_V,R} & =  \sum_{i=0}^{N-1} \left\{ \mathbb{E}_{{\bf x}_{i:i+T_c} \sim p ({\bf x}_{T_c})} \log [D({\bf x}_{i:i+T_c})] \right. \\
     & +  \mathbb{E}_{\widehat{\bf x}_{i:i+T_c} \sim p (\widehat{\bf x}_{T_c})} \log [1-D(\widehat{\bf x}_{i:i+T_c})]   \\
     &   +  \mathbb{E}_{\widehat{\bf x}_{(i+r):(i+r+T_c)} \sim p (\widehat{\bf x}_{T_c})} \log [1 -   \\
     & \left. \hspace*{2.5cm} - D(\widehat{\bf x}_{(i+r):(i+r+T_c)})]  \right\} ,
     \end{aligned}
     \label{Loss3LDV}
\end{equation}
where ${\bf x}_{i:i+T_c} = \{ {\bf x}_{i,j} \}_{j=1}^{T_c}$ and $\widehat{\bf x}_{i:i+T_c} = \{ \widehat{\bf x}_{i,j} \}_{j=1}^{T_c}$, $i=0,\ldots,N-1$ represent the real segmented videos and their reconstructions of $T_c$ frames length, while $p ({\bf x}_{T_c})$ and $ p (\widehat{\bf x}_{T_c})$ are their probabilities. By using the discriminator $L_{D_V,R}$ for reconstructing the term of $\widehat{\bf x}_{(i+r):(i+r+T_c)}$ enforces the generator to learn the temporal relationships of output clips between consecutive states, thus enabling the generation of long-term videos. 
When considering $r=T_c/2$, each video clip, which is part of the longer generated video, overlaps completely with its neighboring video clip segments, half with the previous one and half with the next one, thus ensuring the continuity and consistency in the resulting generated long-temporal video. 

During the training, first the Discriminator is updated by optimizing the content $L_{D_I}$ and movement $L_{D_V}$ using equations~\eqref{LossLDI2} and \eqref{Loss3LDV}, then the Encoder and Generator $L_{REncG}$ are optimized together according to Eq.~\eqref{LossREncG}.

\section{Experimental results}
\label{sec-exps}

In this section, we initially perform a series of experiments using the proposed EncGAN3 model for generating short video clips of 16 frames, and then in the second part we provide results when REncGAN3 is used to generate long-term video sequences of hundreds of frames based on the recall mechanism.
EncGAN3 and the REncGAN3 are implemented using the ADAM optimizer \cite{Adam} with the exponential decay rate of first-order and second-order moment estimation of $\beta_1$=0.5 and $\beta_2$=0.999, while considering a learning rate of $2e^{-4}$ for training all modules: Discriminator, Encoder and Generator. For training the EncGAN3 we consider the same hyper-parameter initialization as for G$^3$AN in \cite{wang_g3an_2020} while using a single V100 GPU with 32 GB memory running on the Ubuntu operating system.

\begin{figure*}[h]
    \centering
        \begin{tabular}{c}
            \includegraphics[width=1\linewidth]{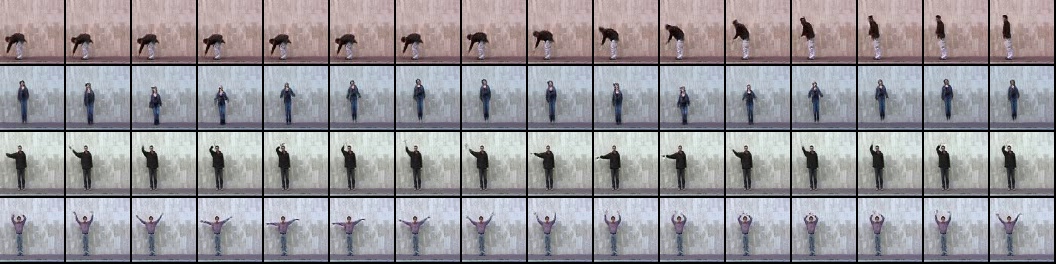} \\
            (a) Frames from the videos generated after training on Weizmann. \\ \includegraphics[width=1\linewidth]{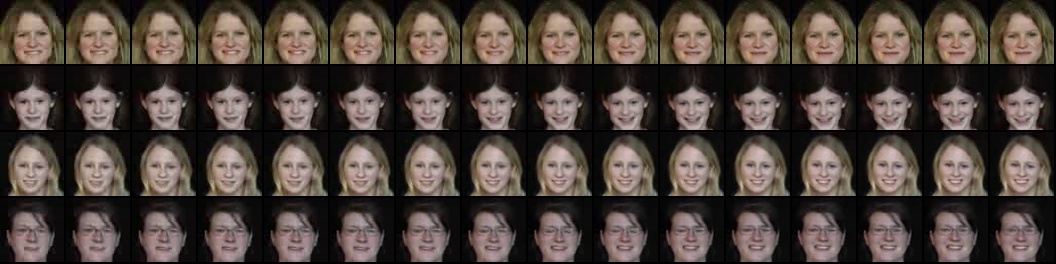} \\
            (b) Frames from the videos generated after training on UvA. \\
        \end{tabular}
    \caption{Generated video frames of $64 \times 64$ pixels resolution by EncGAN3.}
    \label{fig:weiz_uva64_all}
\end{figure*}

\begin{figure*}[h]
    \centering
        \begin{tabular}{cccc}
            \includegraphics[width=0.22\linewidth]{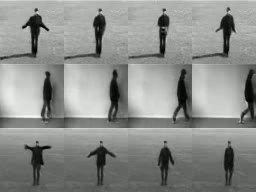} 
            & \includegraphics[width=0.22\linewidth]{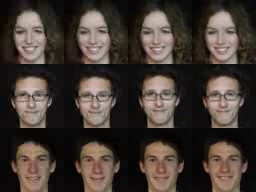}
            & \includegraphics[width=0.22\linewidth]{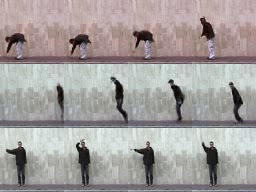}
            & \includegraphics[width=0.22\linewidth]{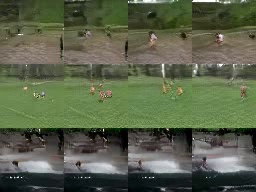} \\
            {\small (a) EncGAN3, KTH} & {\small (b) EncGAN3, UvA} & {\small (c) EncGAN3, Weizmann} & (d) {\small EncGAN3, UCF101} \\
            \includegraphics[width=0.22\linewidth]{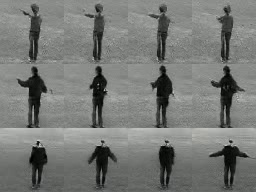} 
            & \includegraphics[width=0.22\linewidth]{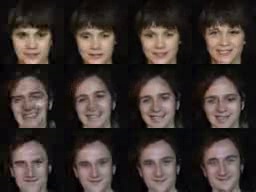}
            & \includegraphics[width=0.22\linewidth]{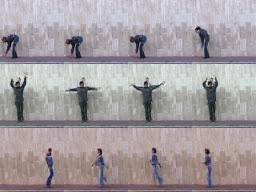}
            & \includegraphics[width=0.22\linewidth]{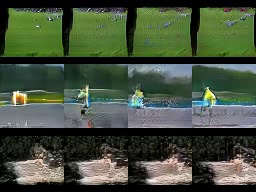} \\
            {\small (e) G$^3$AN, KTH} & {\small (f) G$^3$AN, UvA} & {\small (g) G$^3$AN, Weizmann} & (h) {\small G$^3$AN, UCF101} \\
        \end{tabular}
    \caption{Comparing EncGAN3 ({\it upper panels}) with G$^3$AN ({\it bottom panels}) after being trained on various datasets where the generated videos have a resolution of $64 \times 64$ pixels.}
    \label{fig:64_all_encVSg3}
\end{figure*}

\subsection{Datasets}
\label{sec-datasets}
We consider the following databases for training EncGAN3 for generating videos with 16 frames $(T=15)$ length~: UvA-NEMO~\cite{UvA}, Weizmann~\cite{Weiz}, KTH~\cite{KTH} and UCF101~\cite{UCF101}. UvA dataset contains facial expression movement from 400 individuals. 
The Weizmann dataset comprises 9 subjects performing 10 recorded actions, while KTH contains 25 subjects with six types of actions in four scenes. 
UCF101 includes 101 action categories. For these datasets, we first sample $T=16$ frames as the input video clip from each video, and then crop and resize these to  $64 \times 64$ pixels while preserving the aspect ratios. For the UvA-NEMO dataset, we follow the settings from \cite{wang_g3an_2020} and use their pre-processed $128 \times 128$ pixel resolution dataset. For the UCF101 dataset, we also resize the frames to $128\times 128$ pixels.
In addition, we consider the following databases for training REncGAN3 for generating long videos (hundred+ frames)~: Tai-Chi-HD (Taichi) \cite{Taichi} and Sky-Timelapse (Sky) \cite{Sky}. The Taichi dataset consists of Tai Chi movements, with videos primarily exceeding 100 frames, while the Sky dataset features non-rigid cloud movements with videos of various lengths, ranging from as few as 3 frames to over 3,000 frames. We split each long video into several 16-frame segments, resize them to $128\times 128$ pixels.

\subsection{Evaluation of EncGAN3}

We train the EncGAN3 model, according to the methodology described in Section~\ref{sec-loss_enc}. First, we optimize the Discriminators' loss functions $L_{D_I}$ and $L_{D_V}$ from Eq.~(\ref{LossLDI}) from (\ref{LossLDV}), respectively. Then, we optimize $L_{Enc}$ from (\ref{Obj_Enc}) for the Encoder. Thirdly, we run the model again with the optimized Discriminator and Encoder to minimize the Generator loss $L_G$ from Eq.~(\ref{Obj_Gen}).
The training proceeds with the new batch of data for each iteration, optimizing the parameters of the Encoder, Generator and Discriminator until the validation indicates stable results.

\subsubsection{Qualitative Evaluation}

For the qualitative evaluation, frames from generated videos, after training on the Weizmann dataset, show various human activities in Figure~\ref{fig:weiz_uva64_all}(a), while people displaying variations of facial expressions are displayed in 
Figure~\ref{fig:weiz_uva64_all}(b), after training on the UvA dataset.
We compare the visual results of EncGAN3 with G$^3$AN in Figure~\ref{fig:64_all_encVSg3}. In Figures~\ref{fig:64_all_encVSg3}(a), (b), (c) and (d) we show on each row four frames from videos generated by EncGAN3 following the training with KTH, UvA, Weizmann and UCF101 datasets, respectively, while the videos generated by G$^3$AN\footnote{Code is from 
\url{https://github.com/wyhsirius/g3an-project}.}, after training on the same datasets, are provided in Figures~\ref{fig:64_all_encVSg3} (e), (f), (g) and (h).
The video frames generated by EncGAN3 entail fewer artifacts with less distortion while displaying smooth movement when compared to the frames generated by G$^3$AN, as it can be observed from the first and second row of results from Figure~\ref{fig:64_all_encVSg3}(a), the second row of Figure~\ref{fig:64_all_encVSg3}(b) and in Figure~\ref{fig:64_all_encVSg3}(c), when compared to the frames generated by G$^3$AN shown underneath.

\begin{table*}[h]
    \centering
    \caption{FID results, $\downarrow$ indicates that lower value is better.  indicates that the results are referred from \cite{wang_g3an_2020}.} 
    \vspace*{0.3cm}
    \begin{tabular}{l|ccccc}
        \hline 
            & UvA & Weizmann & KTH & UCF101 \\
            & FID$\downarrow$ & FID$\downarrow$ & FID$\downarrow$ & FID$\downarrow$ \\
        \hline \hline
        VGAN* \cite{vondrick2016vgan} & 235.01 & 158.04 & - & 115.06 \\
        TGAN* \cite{saito2017tgan} & 216.41 & 99.85 & - & 110.58 \\
        MoCoGAN* \cite{tulyakov_mocogan_2018} & 197.32 & 92.18 & - & 104.14 \\
        ImaGINator \cite{wang_imaginator_2020} & -  & 99.80 & - & - \\
        G$^3$AN \cite{wang_g3an_2020} & 91.77 & 98.27 & 111.99 & 108.36 \\
        {\bf EncGAN3}         & {\bf 86.21} & {\bf 83.35} & {\bf 72.59} & {\bf 91.18} \\
    \end{tabular}
    \label{tab-FID_Short}
     \vspace*{-0.3cm}
\end{table*}

\begin{table*}[h]
    \centering
    \caption{Results for IS and its components, where $\uparrow$ means that higher value is better.} 
    \label{tab-IS_Short}
     \vspace*{0.3cm}
    \begin{tabular}{l|ccc|c}
        \hline
            & IS$\uparrow$ & Inter- & Intra- & Dataset \\
            &              &  Entropy $\uparrow$ &  Entropy $\downarrow$ & \\
        \hline \hline
                & 85.44 & 6.041 & 1.593 & UvA \\
        G$^3$AN & 25.54 & 3.924 & 0.684 & Weizmann \\
                & 24.19 & 4.538 & 1.352 & KTH \\
                & 30.01 & 6.903 & 3.501 & UCF101 \\\hline
                & 571.29 & 6.499 & 0.151 & UvA \\
        EncGAN3 & 42.60 & 3.959 & 0.207 & Weizmann \\
                & 50.48 & 4.812 & 0.891 & KTH \\
                & 33.87 & 6.699 & 3.177 & UCF101 \\ 
    \end{tabular}
\end{table*}

\begin{table*}[h]
    \centering
    \caption{Results for FID and IS on UCF101 dataset at a resolution of $128\times 128$ pixels per frame. Results of the other methods are from \cite{mei2023vidm128f_diffu}.}
    \vspace*{0.3cm}
   
        \begin{tabular}{c|cccccc}
         \hline
                &   DVD-GAN & MoCoGAN-HD & DIGAN & StyleGAN-V & VIDM & EncGAN3 \\
                &   \cite{clark2019dvdgan} & \cite{tian2021mocoganhd} & \cite{digan2022iclr} & \cite{stylegan_v2022cvpr} & \cite{mei2023vidm128f_diffu} &  \\
                \hline
            IS($\uparrow$)   & 27.38 & 32.36 & 32.70 & 32.70 & 64.17 & 43.65 \\
            FID($\downarrow$)  & -  & 838 & 577 & - & 263 & 356 \\
        \end{tabular}
     \label{tab:128_Short}
\end{table*}

\subsubsection{Quantitative Evaluation}

For the quantitative evaluation, we adopt the Fréchet Inception Distance (FID) \cite{FID} and Inception Score (IS) \cite{NIPS2016ISimg} adapted for video analysis\footnote{Specifically, video FID \cite{wang_g3an_2020} and video IS \cite{2018VideoVAE_entropy}, hereafter referred to as FID and IS for simplicity.}. Lower FID values\footnote{Evaluation code: \url{https://github.com/wyhsirius/g3an-project/tree/master/evaluation}.} indicate superior visual quality and spatiotemporal consistency, while higher IS values\footnote{Evaluation code: \url{https://github.com/sunxm2357/TwoStreamVAN/tree/master/classifier}.} \cite{sun_twostreamvan_2020} reflect greater diversity in generated videos. Additionally, we evaluate Inter-Entropy and Intra-Entropy \cite{2018VideoVAE_entropy}, components of IS, to separately assess visual quality and diversity.
Table~\ref{tab-FID_Short} compares FID scores for EncGAN3 against G$^3$AN \cite{wang_g3an_2020}, VGAN \cite{vondrick2016vgan}, TGAN \cite{saito2017tgan}, and MoCoGAN \cite{tulyakov_mocogan_2018}. EncGAN3 achieves the lowest FID across all datasets, demonstrating its superior visual quality and spatiotemporal consistency. Table~\ref{tab-IS_Short} presents IS, Inter-Entropy, and Intra-Entropy results, where higher Inter-Entropy indicates better diversity and lower Intra-Entropy signifies improved visual quality. EncGAN3 provides the best results on the Weizmann, KTH, and UvA datasets for these metrics.
In Table \ref{tab:128_Short}, we compare FID and IS on the UCF101 dataset at $128\times 128$ resolution, pitting EncGAN3 against DVD-GAN \cite{clark2019dvdgan}, MoCoGAN-HD \cite{tian2021mocoganhd}, DIGAN \cite{digan2022iclr}, StyleGAN-V \cite{stylegan_v2022cvpr}, and Video
Implicit Diffusion Models (VIDM) \cite{mei2023vidm128f_diffu}. EncGAN3 surpasses all GAN-based methods in both metrics. While VIDM, based on a diffusion architecture, performs well for short video generation at a significantly larger computational cost, our method REncGAN demonstrates superior results in long-term video generation, as shown in Table \ref{tab:FVD_R2vs}, highlighting its advanced temporal processing capabilities.

\begin{figure*}[h]
    \centering
        \begin{tabular}{cc}
        \hspace*{-0.5cm}
            \includegraphics[width=0.53\linewidth]{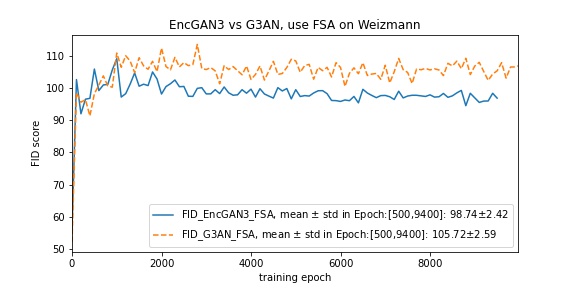} \hspace*{-0.3cm} &  \hspace*{-0.3cm}\includegraphics[width=0.53\linewidth]{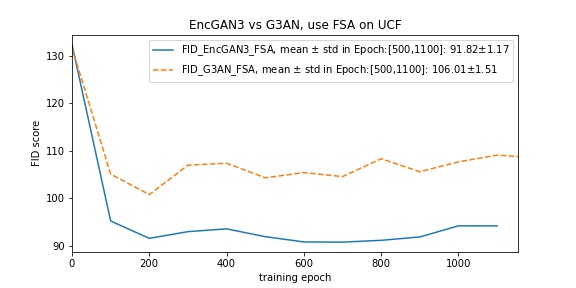}  \\
            (a) Weizmann \hspace*{-0.3cm} &  \hspace*{-0.3cm} (b) UCF101 \\
        \end{tabular}
    \caption{The convergence during the training, indicated by the FID scores evaluated for every 100 epochs, for EncGAN3 ({\it solid lines}) and G$^3$AN ({\it dashed lines}) when training on Weizmann and UCF101 datasets.}
    \label{fig:plotFID}
\end{figure*}

In order to assess the convergence, we evaluate the FID during the entire training of the EncGAN3 on Weizmann and UCF101 in Figures~\ref{fig:plotFID}(a) and \ref{fig:plotFID}(b), respectively. We observe that EncGAN3 has better FID scores than G$^3$AN on Weizmann and UCF101 datasets. 
The FID results for the initial training epochs are sometimes not reliable, as it can be observed from Figure~\ref{fig:plotFID}(a), indicating that the videos generated at those stages contain sometimes random artifacts which are not consistent with real videos.

\begin{figure*}[h]
    \centering 
    \hspace*{-0.5cm}
        \begin{tabular}{ccc}
            \includegraphics[width=0.33\linewidth]{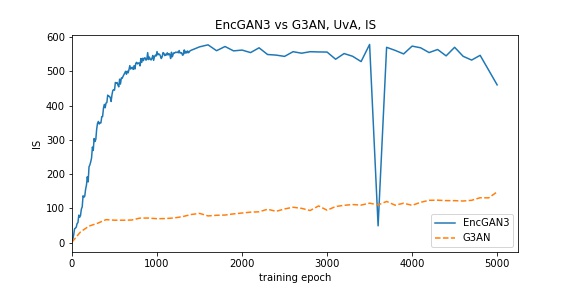}  & 
            \includegraphics[width=0.33\linewidth]{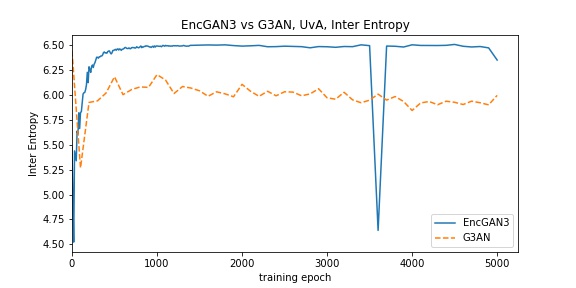}  & \includegraphics[width=0.33\linewidth]{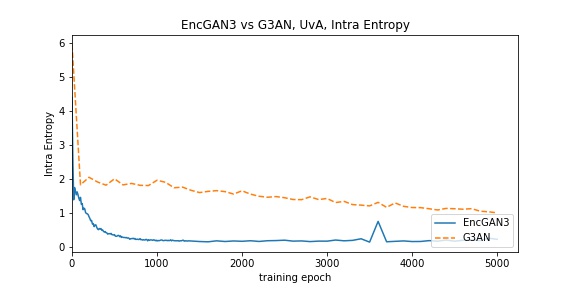} \\
            (a) IS for UvA   &  (b) Inter-entropy  for UvA  &  (c) Intra-entropy for UvA \\
            \includegraphics[width=0.33\linewidth]{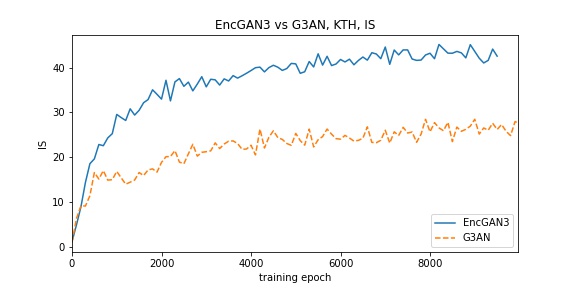}  &  \includegraphics[width=0.33\linewidth]{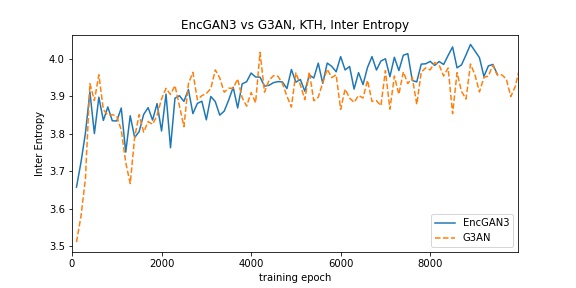}  &   \includegraphics[width=0.33\linewidth]{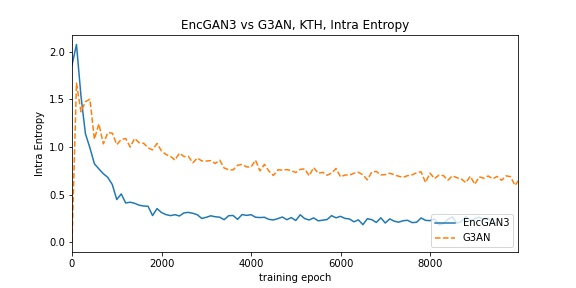} \\
            (d) IS for KTH   &  (e) Inter-entropy  for KTH  &  (f) Intra-entropy for KTH \\
        \end{tabular}
    \caption{Evaluating Inception Score (IS) and its components inter-entropy and intra-entropy of EncGAN3 ({\it solid line}) and G$^3$AN ({\it dash line}) when trained on UvA and KTH datasets.}
    \label{fig:plotIS}
\end{figure*}

We also plot the IS and its components inter-entropy and intra-entropy for the UvA database in Figures~\ref{fig:plotIS}(a), (b) and (c), respectively, while the same measures for the KTH database are provided in Figures~\ref{fig:plotIS}(d), (e) and (f), respectively.
Higher IS means better visual quality and diversity while higher inter-entropy indicates better diversity and lower intra-entropy corresponds to better visual quality. 
EncGAN3 produces videos of a similar diversity to G$^3$AN when training on the KTH dataset, as shown in Figure~\ref{fig:plotIS}(e) but with better visual quality as shown by the results from Figure~\ref{fig:plotIS}(f). It indicates that the encoder constrains the generator to produce videos with better visual characteristics while not limiting their diversity when training on the human actions KTH dataset. Meanwhile, the encoder improves both the visual quality as well as the diversity of generated videos when trained on the face expression UvA dataset, according to the results from Figures~\ref{fig:plotIS}(b) and \ref{fig:plotIS}(c).

\subsubsection{Ablation Studies}
\label{sec-ablation_enc}

In the following, we test the contribution of each component of the architecture from Figure~\ref{fig:EncGAN3}, using different sampling strategies for the learning set as well as when considering different loss functions or hyper-parameters for the EncGAN3 training.

\begin{table*}[h]
    \centering
    \caption{Testing the contribution of the EncGAN3 architecture components.}
    \label{tab-encG}
    \vspace*{0.3cm}
    \begin{tabular}{l|cc|cc|cc}
    \hline
    Architecture & \multicolumn{2}{c|}{UvA} & \multicolumn{2}{c|}{Weizmann} & \multicolumn{2}{c}{KTH} 
  \\ 
    & FID$\downarrow$ & IS$\uparrow$ & FID$\downarrow$ & IS$\uparrow$ & FID$\downarrow$ & IS$\uparrow$ \\
    \hline \hline
    no $G_C$, $G_T$ & 95.500 & 63.926 & 101.638 & 2.244 & 73.220 & 2.867 \\
    no $G_C$ & 88.058 & 133.352 & 89.004 & 7.020 & 75.309 & 3.853 \\ 
    no $G_T$ & 90.713 & 537.852 & 97.554 & 5.564 & 74.963 & 4.966 \\ 
    no F-SA & 87.526 & - & 82.821 & - & 73.792 & - \\
    no Enc & 93.258 & 148.216 & 98.564 & 6.303 & 75.388 & 2.328 \\
    EncGAN3 & 86.210 & 571.29 & 78.935 & 8.906 & 70.448 & 5.986 \\ 
    \end{tabular}
    \vspace*{-0.3cm}
\end{table*}

{\bf EncGAN3 module components}. We test the contribution of the components to the efficiency of the EncGAN3, such as the Encoder and the three-stream processing pipeline, F-SA module in the Generator $G$, considering three datasets for the training, UvA, Weizmann and KTH, with the results provided in Table~\ref{tab-encG}. We consider assessing the results  when removing the Encoder (Enc) or the F-SA processing modules, and also when removing the auxiliary reconstruction streams for the content $G_C$ and movement $G_T$. 
From the results provided in Table~\ref{tab-encG}, we observe that the presence of two auxiliary streams $G_C$ and $G_T$, as well as the Encoder and F-SA module, is important for the performance of EncGAN3.

{\bf Changing the loss function.}
In the following, we study the performance when changing the loss functions described in Section~\ref{sec-loss_enc}, which uses video reconstruction, and consider the reconstruction error of difference maps in $L_{Enc}$ from Eq.~(\ref{Obj_Enc}). The Encoder's loss function, in this case, becomes~:
\begin{align}
    L_{Enc,{\bf v}}  = &  \sum_{i=1}^N \| {\bf x}_{i0} - \widehat{\bf x}_{i0} \|^2  + \sum_{i=1}^N  \sum_{j=1}^T \| {\bf v}_{ij} - \widehat{\bf v}_{ij} \|^2 \notag \\
    & - D_{KL} (q_{{\theta}_{\bf x}} ({\bf z}_{\bf x}\mid{\bf x})) \| p ({\bf z}_{\bf x}) )  \notag \\
    & - D_{KL} (q_{{\theta}_{\bf v}} ({\bf z}_{\bf v}\mid{\bf v})) \| p ({\bf z}_{\bf v}) ), \label{Obj_Enc1}
\end{align}
where the second term represents the error in the difference maps in the sequence of $T$ frames and $N$ video sequences.

\begin{table}[h]
    \caption{Results after changing the loss functions and learning rate (L-rate).}
    \vspace*{0.2cm}
    \centering
    \begin{tabular}{l|cc|c|cc}
    \hline
    &  \multicolumn{2}{c|}{Baseline} & \multicolumn{1}{c|}{$L_{Enc,{\bf v}}$} & \multicolumn{2}{c}{$L_{Enc,{\bf v}}$+$L_{G,{\bf v}}$}
    \\   \hline
    L-rate  & 2e-4 & 4e-5 & 2e-4 & 2e-4 & 4e-5 \\ \hline
    FID$\downarrow$ & {\bf 87.76} & 88.68 & 90.77 & 95.02 & 89.71 \\
    \end{tabular}
    \label{tab:loss_diff}
\end{table}

We also consider changing the reconstruction term only in the Encoder's loss $L_{Enc,{\bf v}}$, as in Eq.~(\ref{Obj_Enc1}), or in both encoder and generator, marked as $L_{Enc,{\bf v}}+L_{G,{\bf v}}$. Besides all these changes, we also consider decreasing the learning rate from $2e^{-4}$ to $4e^{-5}$ when training all modules. Meanwhile, the loss function from Eq.~\eqref{Obj_Enc} is considered as the baseline. According to the results from Table~\ref{tab:loss_diff} for $L_{Enc,{\bf v}}+L_{G,{\bf v}}$, we observe that we achieve the best results by decreasing the learning rate five times for the loss function when considering the movement error for both Encoder and Generator. These results indicate that it is important to consider the full frame reconstruction in the loss function instead of Eq.~\eqref{Obj_Enc1}. Loss functions such as $L_{Enc}$ from Eq.~(\ref{Obj_Enc}) and $L_G$ from 
Eq.~(\ref{Obj_Gen}) do not only optimize the movement stream reconstruction, but also consider how the movement representation is employed to reconstruct realistic video streams, frame after frame.

\begin{figure*}[h]
    \centering
        \begin{tabular}{cc}
            \includegraphics[width=0.46\linewidth]{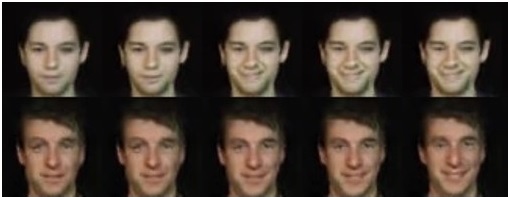} & \hspace{6.5mm}
            \includegraphics[width=0.46\linewidth]{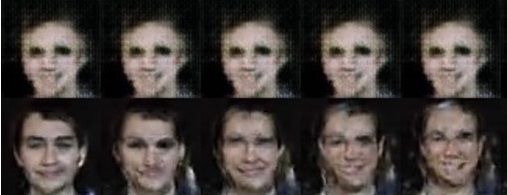} \\
            {\small (a) EncGAN3, uniform sampling} &
            {\small (b) G$^3$AN, uniform sampling} \\
            \includegraphics[width=0.46\linewidth]{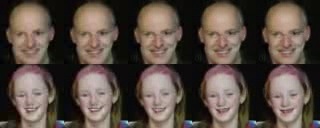} & \hspace{6.5mm}
            \includegraphics[width=0.46\linewidth]{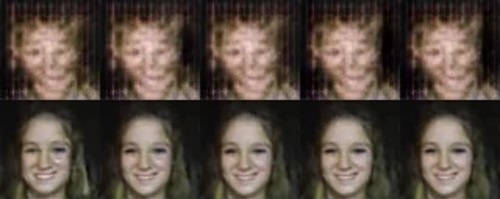} \\
            {\small (c) EncGAN3, step sampling} &
            {\small (d) G$^3$AN, step sampling} \\
        \end{tabular}
    \caption{Generated frames for EncGAN3 in (a), (c) and G$^3$AN in (b), (d) when using uniformly or step sampled training sets trained for 100 ({\it top row}) and 5000 epochs ({\it bottom row}).}
    \label{fig:uni_enc_g3}
\end{figure*}

{\bf Video-sampling strategies for training EncGAN3.}
We provide the results when considering two different sampling strategies when training EncGAN3 on the UvA database~: step sampling and uniform sampling. The step sampling used in G$^3$AN \cite{wang_g3an_2020}, consists of randomly selecting a starting frame and then sampling the following video frames with a sampling step of 2 or 3 for the entire number of frames considered. Meanwhile, we consider the uniform sampling that divides a video clip into 16 sets with equal numbers of frames, and then  randomly sample one of the frames from within each set. 
The uniform sampling represents the temporal information covering an entire video sequence. The uniform sampling models better the variation of the information over the entire video sequence when compared to the step sampling.
When considering uniform sampling, the frames from videos generated by EncGAN3 trained on the uniformly sampled training set, after 100 epochs, and after 5000 epochs, are provided on the top and bottom row of Figure~\ref{fig:uni_enc_g3}(a), respectively. Meanwhile, frames from videos generated by G$^3$AN, after being trained on the uniformly sampled training set for 100 and 5000 epochs are shown at the top and bottom rows of Figure~\ref{fig:uni_enc_g3}(b), respectively. 
From these results based on the more complex uniformly sampled training set, it can be observed that EncGAN3 provides significantly better results than G$^3$AN given that by using an Encoder for inference leads to more stable modeling, quicker convergence and better generation results.

As it can be observed from the results in Figure~\ref{fig:uni_enc_g3}, EncGAN3 provides worse results when considering the uniform sampling of the training set than the results for EncGAN3 and G$^3$AN using the step sampling approach for the training set. However, EncGAN3 trained on the step sampled training set provides good results after about 3000 epochs and G$^3$AN after about 5000 epochs. EncGAN3 requires about 20\% training time more than G$^3$AN for each epoch. However, overall in order for EncGAN3 to achieve the convergence after 3000 epochs of training it requires much less training time than G$^3$AN, which needs 5000 epochs for convergence. Meanwhile, EncGAN3 trained on the uniformly sampled training set converged after about 100 epochs while generating frames displaying better movement than those generated by using the step sampling strategy at the  same epoch, indicating that by training using uniformly sampling frames as the training set requires much less training time. Thus, the best strategy for training EncGAN3 is by initially training on uniformly sampled frames and then after 200-300 training epochs adopting the step sampled frames, which can lead to learning richer temporal information while reducing the training time.

\begin{figure*}[h]
    \centering
        \begin{tabular}{cc}
            \includegraphics[width=0.46\linewidth]{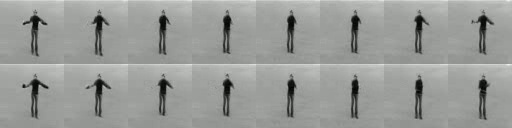} 
            & \includegraphics[width=0.46\linewidth]{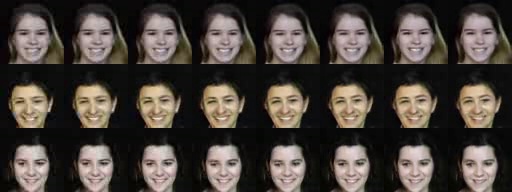} \\
            {\small (a) ${\bf z}_{c_1}$,${\bf z}_{m_1}$ ({\it top}) and ${\bf z}_{c_1}$,${\bf z}_{m_2}$ ({\it bottom})} &
            {\small (b) ${\bf z}_{c_3}$,${\bf z}_{m_3}$ ({\it top}), ${\bf z}_{c_4}$,${\bf z}_{m_4}$ ({\it middle})}\\
             & {\small and their sum ({\it bottom}) \quad} \\
        \end{tabular}
    \caption{Frames generated from the manipulated latent codes.}
    \label{fig:latent_vids}
\end{figure*}

{\bf Video generation results following latent code manipulation.} 
To explore the relationship between latent codes and generated frames, we progressively tune the latent codes and observe the changes in the generated video clips, as displayed in Figure~\ref{fig:latent_vids}. 
We firstly fix the content latent code ${\bf z}_{c_1}$ while we consider different motion latent codes 
${\bf z}_{m_1}$, ${\bf z}_{m_2} $ as inputs for the Generator with the video results shown on top and bottom of Figure~\ref{fig:latent_vids}(a). We can observe that the generated frames show the same subject clapping hands while having different particular movements. Meanwhile, in Figure~\ref{fig:latent_vids}(b), on the top and middle rows we show video frames generated when considering 
$\{ {\bf z}_{c_3}$, ${\bf z}_{m_3} \} $ and $\{ {\bf z}_{c_4}$, ${\bf z}_{m_4} \} $, while in the bottom row we show frames generated when considering the sum of the latent variables corresponding to content and movement
$ \{{\bf z}_{c_3}+{\bf z}_{c_4}, {\bf z}_{m_3}+ {\bf z}_{m_4} \}$.
We observe that the frames generated using the sum of the latent codes, corresponding to the other two video sequences, inherit and combine some of their characteristics.

\begin{figure*}[h]
    \centering
    \hspace*{-0.3cm}
    \scalebox{0.8}{
        \begin{tabular}{cccc}
            \includegraphics[width=0.29\linewidth]{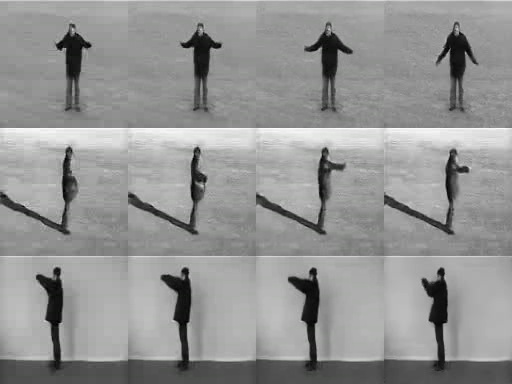} 
            & \includegraphics[width=0.29\linewidth]{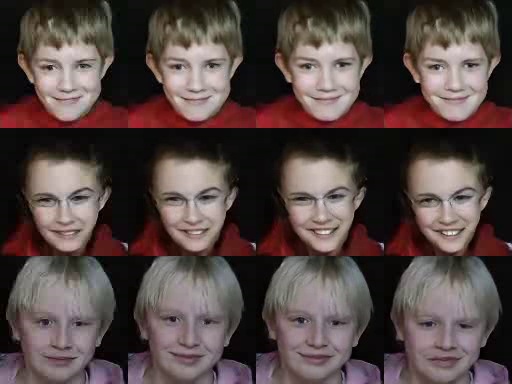}
            & \includegraphics[width=0.29\linewidth]{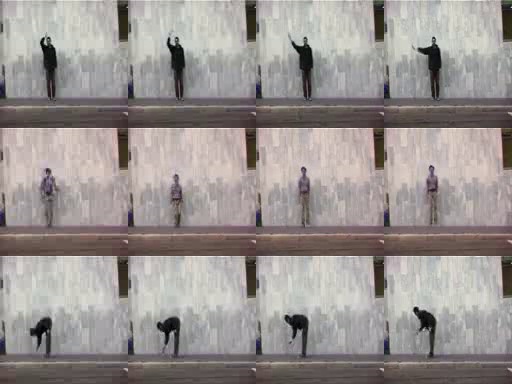}
            & \includegraphics[width=0.29\linewidth]{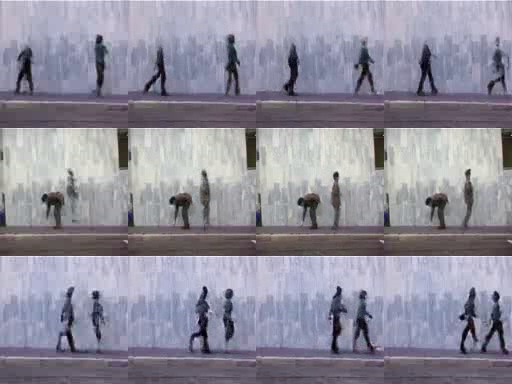} \\
            {\small \bf (a) KTH} & {\small \bf (b) UvA} & {\small \bf (c) Weizmann} & {\small \bf (d) Two Objects} \\
        \end{tabular}
    }
    \caption{Generated video frames of EncGAN3 on KTH (a), UvA (b) and Weizmann (c, d) datasets in the resolution of $128 \times 128$ pixels. 
    Video data corresponding to these frames and additional video sequences are provided in the Online Resource from the supplementary material files attached to this paper.
    }
    \vspace*{-0.3cm}
    \label{fig:128x128}
\end{figure*}

{\bf Generation of higher resolution and complexity videos.}
We use EncGAN3 for generating videos of $128 \times 128$ pixels resolution by adding an extra $G^3_5$ module into the Generator and other extra convolution and fully-connected layers for the Discriminator and Encoder in the architecture from Figure~\ref{fig:EncGAN3}. The generated frames are displayed in 
Figures~\ref{fig:128x128}(a), (b) and (c) after training on KTH, UvA and Weizmann datasets. 
Moreover, we observe that when training on the Weizmann dataset for many epochs, such as 9,900 - 10,000 epochs, EncGAN3 could generate videos containing two moving objects, as shown in Figure~\ref{fig:128x128}(d). 
This can happen because of the differences in the feature processing between Encoder and Generator caused when adding different types of extra layers.\footnote{EncGAN3 generates videos of higher $128\times 128$ pixel resolution from the original resolution of $64\times 64$ pixels by adding extra layers. We added fully connected (fc) instead of conv layers in $Enc$ while deconv and conv layers in $G$ and $D$, respectively.} After learning to model a single moving object, the Generator combines the latent codes, corresponding to different video scenes, and then reconstructs scenes showing two moving objects.
The FID scores for the model trained on the KTH, UvA and Weizmann datasets in the resolution of $128 \times 128$ are 75.28, 97.8 and 86.42, respectively. 

\subsubsection{Memory requirements for the video generation}

EncGAN3 is trained on 16-frame videos. Simply extending the training video length to capture longer sequences would cause a steep rise in GPU memory requirements. As detailed in Table \ref{tab:memory_cost}, increasing the spatial resolution of the generated videos by EncGAN3 leads to a relatively moderate increase in the memory usage, resulting in having to reduce the batch size from 10 to 1 in order to be able to generate videos of $128 \times 128$ instead of $64 \times 64$. In contrast, even a small increase in the generated video length, such as from 16 to 20 generated video frames, results in an exponential rise in the memory demand leading to out-of-memory for the CUDA operating system on a V100 GPU with 32 GB memory, highlighting the inefficiency of temporal representation compared to spatial representation. This inefficiency motivates further research into developing a more efficient video representation mechanism for the temporal dimension, which can preserve the ability to model diverse types of video data while substantially reducing the computational requirements for generating such dense temporal representations.

\begin{table}[h]
    \centering
    \begin{tabular}{cc|c}
       \hline
        Video length    &   Resolution  &   Batch size      \\
        (frames)        &   (pixels)    &   (16-frame videos) \\  \hline\hline
         16     &   $64\times64$    &   10  \\
         16     &   $128\times128$  &   1   \\
         20     &   $64\times64$    &   out-of-memory  \\
    \end{tabular}
    \caption{Memory requirements for the EncGAN3 when training on a V100 GPU with a memory of 32GB.}
    \label{tab:memory_cost}
\end{table}

These experiments highlight the challenges in generating long-term videos, providing the motivation for the development of the recall mechanism, which is introduced in Section~\ref{sec-recall}. 
It is important to note that while the recall mechanism does not directly improve the underlying algorithm's memory utilization efficiency or alter its storage structure, it effectively mitigates the GPU memory bottleneck by decomposing the video generation process into memory-efficient sub-sequences. By maintaining a fixed memory footprint regardless of the overall video length, the recall mechanism enables the generation of long videos—up to hundreds of frames—with improved temporal dynamics and high visual quality without increasing the training video length.

Other video generation approaches that have low memory requirements have been using simple interpolations between consecutive frames, \cite{digan2022iclr,stylegan_v2022cvpr}. For instance, DIGAN \cite{digan2022iclr} attempts to generate longer videos by training on short sequences with low frame rates and extending their length through interpolation. However, DIGAN struggles to generate videos that exceed 100 frames, underscoring the need for models that can effectively learn temporal dynamics over longer durations to generate coherent extended sequences. Moreover, the temporal interpolation does not add any new information leading to the diversification in the longer videos, but rather it makes them temporally smoother.

\subsection{Long-temporal generated video sequences results}  

In this section, we present and analyze the long-temporal video generation results, when using REncGAN3, described in Section~\ref{TrainRecall}, which employs the recall mechanism, for generating video sequences of hundreds of frames.
We consider training on the Tai-Chi-HD~\cite{Taichi} and Sky-Timelapse~\cite{Sky} databases for generating videos with over a hundred frames by following the long-term video generation methodology.
Taichi-HD (Taichi), contains 264 in-the-wild Tai-Chi videos from YouTube showing different practitioners performing complex movement sequences with different backgrounds, while the Sky-timelapse (Sky) dataset contains videos displaying movements of clouds in the sky, under various lighting conditions.
Taichi dataset contains just 4 videos of lengths over 1024 frames while Sky contains more than 200 such long-term videos. 
REncGAN3 is implemented using the same settings as EncGAN3, which are described in the first paragraph from Section~\ref{sec-exps}.
The generation of $128 \times 128$ pixels videos trained on Taichi and Sky datasets is created using one A60 GPU with 46GB memory using the Ubuntu operating system.

\subsubsection{Qualitative results for the long-temporal generated videos}
 
\begin{figure*}[h]
    \centering
    \scalebox{0.33}{
        \begin{tabular}{ccc}
            \includegraphics[width=1\linewidth]{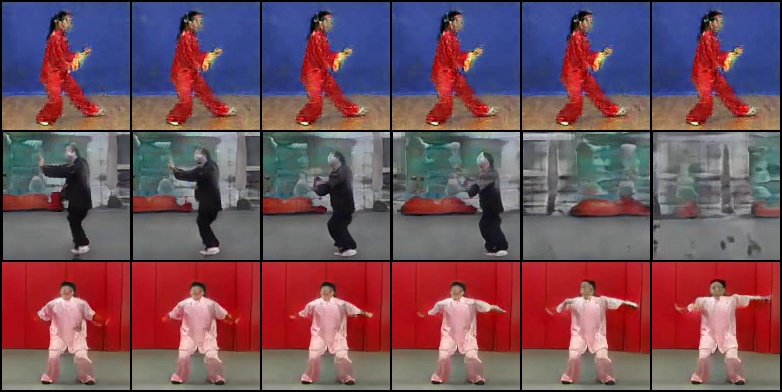} & \includegraphics[width=1\linewidth]{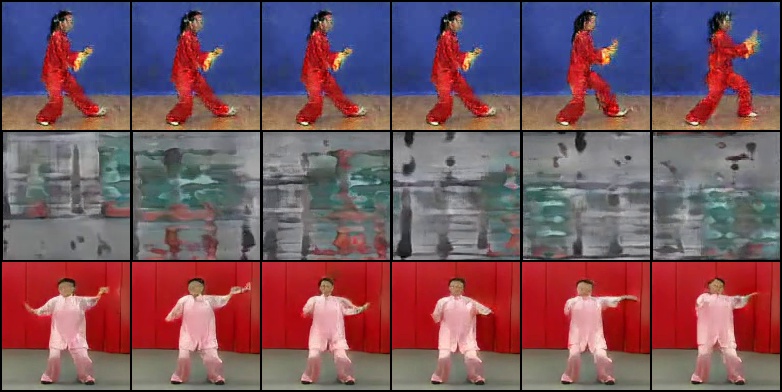} & \includegraphics[width=1\linewidth]{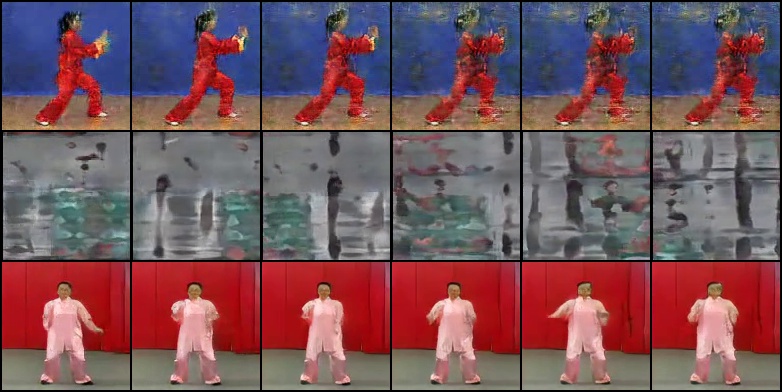} \\
        \end{tabular}
    }
    \caption{Each row from top to bottom shows frames from videos generated by REncGAN3, DIGAN~\cite{digan2022iclr} and TATS~\cite{ge2022tatsLong}, respectively. The generated video lengths are of 388, 424, and 1024 frames for videos in each row, where the frames are of resolution $128 \times 128$ pixels and are sampled with a step of 8 from frame sequences between 0 to 130 frames (left), between 130 to 260 frames (middle) and between 260 to 400 frames (right). 
    Corresponding and additional video data are given in online resources provided in the supplementary material files associated with this paper. 
    }
    \label{fig:longlen_taichi}
\end{figure*}

\begin{figure*}[t]
    \centering
    \scalebox{0.33}{
        \begin{tabular}{ccc}
            \includegraphics[width=1\linewidth]{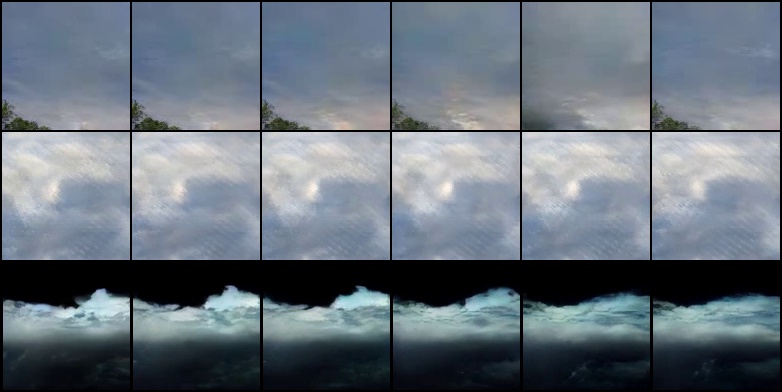} & \includegraphics[width=1\linewidth]{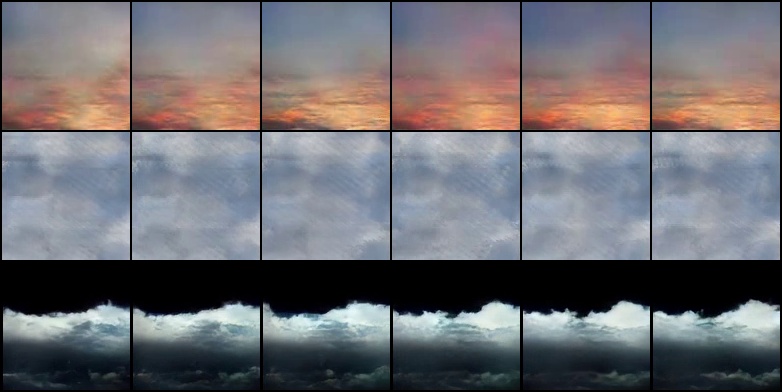} & \includegraphics[width=1\linewidth]{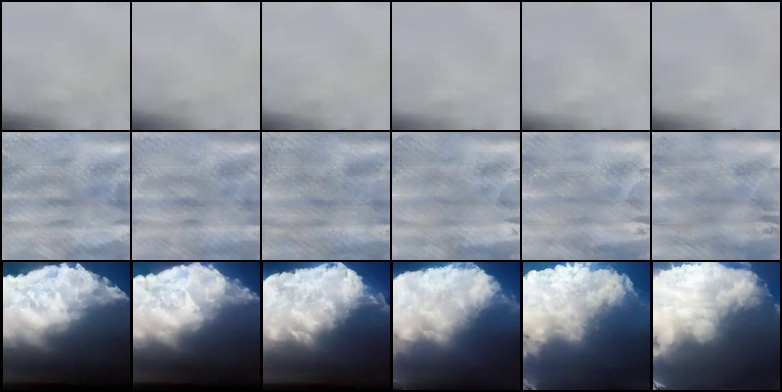} \\
        \end{tabular}
    }
    \caption{Each row from top to bottom shows frames from videos generated by REncGAN3, DIGAN~\cite{digan2022iclr} and TATS~\cite{ge2022tatsLong}, respectively. The generated video lengths are 1324, 1024 and 1024 frames for videos in each row, where the frames are of resolution $128\times 128$ pixels, sampled with a step of 8 frames from video sequences of lengths 0-300 (left), 300-600 (middle) and 600-900 (right), frames.}
    \label{fig:longlen_sky}
\end{figure*}

Following the implementation of REncGAN3, we generate long-temporal video sequences by connecting pairs of video clips. For the long-term video generation, we consider 50\% overlapping between the consecutive video clips and consequently, we do not need the second half from the first video clip of each pair. Figures~\ref{fig:longlen_taichi} and~\ref{fig:longlen_sky} show generated videos containing frames of $128\times 128$ pixels, generated by REncGAN3, and compare them with the results provided by DIGAN~\cite{digan2022iclr} and TATS~\cite{ge2022tatsLong} on Taichi and Sky datasets, respectively. From the first row of Figure~\ref{fig:longlen_taichi} we can observe that the frames generated by REncGAN3 show temporal consistency and continuity, while DIGAN cannot maintain the consistency in the generated frames, displaying rather blurred features, while TATS shows repeated movements which are not consistent with the Taichi action. Meanwhile, the frames generated by REncGAN3 show slow Taichi movements which is fitting to the movement from the original videos, while the other methods generate videos displaying rather quick movements which do not correspond to the Taichi movement from videos. 

\begin{figure*}[h]
    \centering
    \scalebox{1}{
        \begin{tabular}{c}
            \includegraphics[width=1\linewidth]{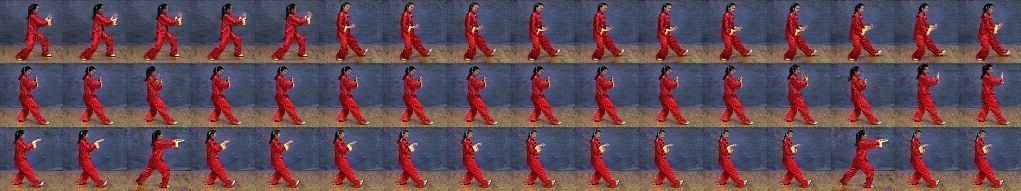} \\
            {\small (a) Generated from the mean value of latent spaces.} \\ 
            \includegraphics[width=1\linewidth]{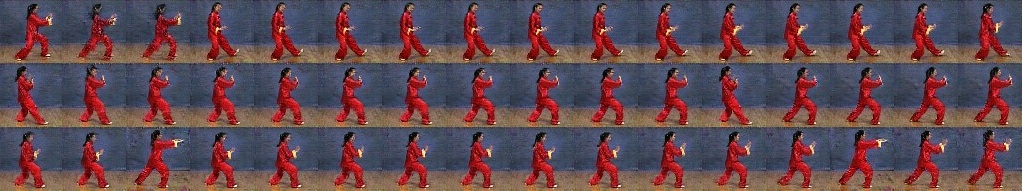} \\
            {\small (b) Generated from sampled latent codes.} \\
        \end{tabular}
    }
    \caption{Frames sampled from generated long-term video sequences.}
    \label{fig:recall_longlen}
\end{figure*}

In  Figure~\ref{fig:longlen_sky}, the frames generated by REncGAN3 show clouds gradually covering the trees in the left panel and with less noise than those in the second row which also display some unrealistic artifacts.
Moreover, REncGAN3 is able to generate videos of various lengths without using interpolation or extrapolation, as DIGAN does in~\cite{digan2022iclr}, while it relies on a simple recall mechanism, which connects short-term clips, as proposed in this paper. In addition, as shown in Figure~\ref{fig:recall_longlen}, although generating long videos of the human action Taichi sequences with good motion dynamics is hard~\cite{digan2022iclr} due to the strict constraints of physical body specific to these actions, our REncGAN3 model provides visual results of good quality when generating the video sequence from sampled latent codes, as shown in Figure~\ref{fig:recall_longlen} (b), as well as by using the mean value, as shown in Figure~\ref{fig:recall_longlen} (a). The obvious motion dynamics benefit from the mixture of the clip-by-clip and frame-by-frame modeling together with the generation processes in the recall mechanism, as proposed in this paper.

\begin{table*}[h]
    \centering
    \caption{Comparison of FVD scores on generated long videos ($128\times128$ resolution). FVD is measured for sub-sequences of 16 and 128 frames, denoted as FVD-16f and FVD-128f, respectively. The ratio of FVD-16f to FVD-128f quantifies the degradation in frame quality over longer durations, reflecting both individual frame quality and temporal coherence. The results of other methods are taken from \cite{mei2023vidm128f_diffu, nips2022dynamicLong} to ensure consistency in resolution and video length. \textbf{Bold} indicates the best results, while \textit{italics} denote the second-best results.}
    \label{tab:FVD_R2vs}
    \vspace*{0.3cm}
        \begin{tabular}{l|ccc|ccc}
        \hline
                    & \multicolumn{3}{c|}{Taichi-128$\times$128}   & \multicolumn{3}{c}{Sky-128$\times$128}  \\\hline
                    & FVD-16f$\downarrow$ & FVD-128f$\downarrow$ & ratio$\uparrow$            
                    & FVD-16f$\downarrow$ & FVD-128f$\downarrow$ & ratio$\uparrow$        \\
        \hline\hline
        MoCoGAN-HD \cite{tian2021mocoganhd}   & 144.7     & -         & -     & 183.6     & 635.6     & 0.28      \\
        DIGAN \cite{digan2022iclr}          & 128.1     & 748.0     & 0.17  & {\it 114.6}     & {\it 228.6}     & 0.50      \\
        StyleGAN-V \cite{stylegan_v2022cvpr}  & 143.5     & 691.1     & 0.2   & -         & -         & -         \\
        TATS \cite{ge2022tatsLong}              & {\bf 94.6}      & -         & -     & 132.5     & 435.0     & 0.30      \\
        Long-Video-GAN \cite{nips2022dynamicLong}   & -         & -         & -     & {\bf 107.5}     & {\bf 142.6}     & {\bf 0.75}      \\
        VIDM \cite{mei2023vidm128f_diffu} & 121.9     & {\it 563.6}     & {\it 0.21}  & -         & -         & -         \\
        \hline
        our REncGAN3  & {\it 113.5}     & {\bf 145.9}     & {\bf 0.77}  & 360.9     & 587.0     & {\it 0.61}      \\   
        \end{tabular}
\end{table*}

\subsubsection{Quantitative results for long-temporal video generation}
\label{QuanrEncGAN3}

In Table~\ref{tab:FVD_R2vs}, we evaluate the quality of videos generated by our REncGAN3 models using the Fr\'echet Video Discriminator (FVD) metric \cite{unterthiner2019fvd}\footnote{Code is available at \url{https://github.com/google-research/google-research/tree/master/frechet_video_distance}.}, specifically FVD-16f and FVD-128f, which measure the first 16 and 128 frames of the generated sequences, respectively. Lower FVD values indicate superior visual quality and spatiotemporal consistency. Additionally, we compute the ratio of FVD-16f to FVD-128f to quantify the degradation in frame quality over longer durations, which reflects both individual frame quality and temporal coherence.
As shown in Table~\ref{tab:FVD_R2vs}, our method achieves the best results for FVD-128f and FVD ratio, along with the second-best result for FVD-16f in generating TaiChi movements. This demonstrates its effectiveness in producing long-duration videos of complex rigid motions. Our method stitches generated clips into longer sequences, ensuring temporal coherence and dynamics. While this approach excels in generating rigid motions like TaiChi, the lack of interpolation in the stitching process can compromise temporal coherence, particularly in non-rigid motions such as cloud movements in the Sky dataset. Nevertheless, our method still achieves the second-best FVD ratio for Sky-related movements, highlighting its robustness against quality degradation in extended frame generation.

\begin{figure}[t]
    \centering
    \includegraphics[width=1\linewidth]{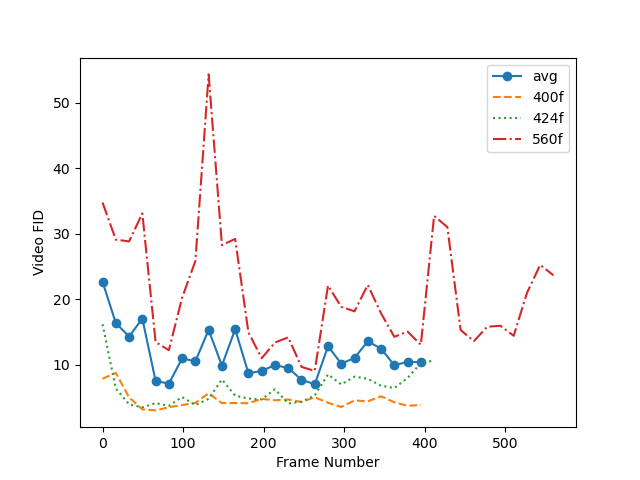}
    \caption{Video FID calculated from non-overlapping 16-frame clips sampled consecutively from long-term videos generated by REncGAN3 trained on TaiChi dataset for lines labeled with `400f' indicate the video length while ``avg'' represents the average results calculated on segment by segment basis from the long generated sequence.
    }
    \label{fig:plot_fid_longlen}
\end{figure}

To further measure the quality degradation, we evaluate the consistency and continuity over the generated long video sequences by segmenting them into consecutive non-overlapping video segments of 16 frames from the long-term generated video, and then we evaluate the FVD and the Fréchet Inception Distance (FID) \cite{FID} in the video domain \cite{wang_g3an_2020}, on each video segment individually. 
Similar to FVD, lower video FID values mean better visual quality and spatial-temporal consistency of generated videos. In this way, we track the evaluation of the entire video's continuity and consistency in the content and movement quality. Such an evaluation approach can be applied to videos of arbitrary lengths. 
In Figure~\ref{fig:plot_fid_longlen}, we provide the video FID values for three long-term generated videos after training on the TaiChi dataset as well as their average. Two of the generated videos display consistency in good FID scores, while the one indicated in red and labeled as 560f, displays more complex variation with some segments characterized by high FID scores.
The recall mechanism in REncGAN3 merges short-term video clips instead of generating them frame by frame, displaying good quality consistency and addressing the degradation problem present in the long-term videos generated by other methods, as shown by the FID results from Figure~\ref{fig:plot_fid_longlen}.


\begin{figure*}[t]
    \centering
    \begin{tabular}{cc}
        \includegraphics[width=0.5\linewidth]{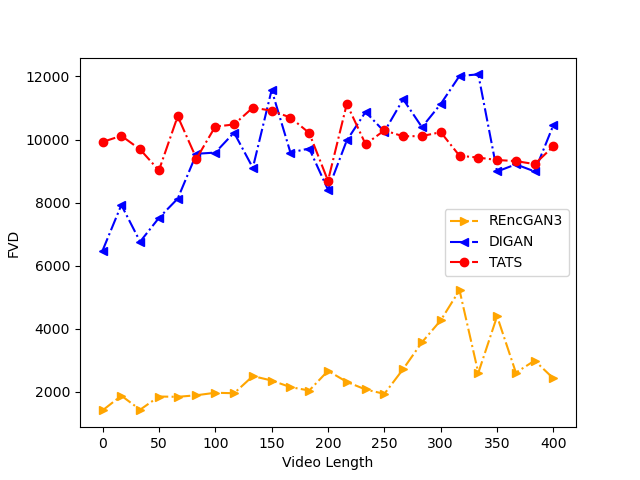} &
        \includegraphics[width=0.5\linewidth]{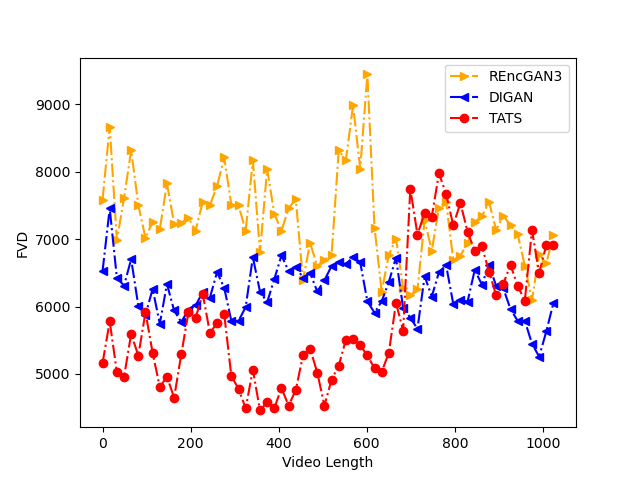} \\
        (a) TaiChi-HD & (b) Sky-Timelapse
    \end{tabular}
    \centering
    \caption{FVD of non-overlapping 16-frame clips sampled from long-term videos generated by REncGAN3, DIGAN~\cite{digan2022iclr} and TATS \cite{ge2022tatsLong} after training on TaiChi (a) and Sky (b) datasets. }
    \label{fig:plot_fvd_longlen_vs}
\end{figure*}

Moreover, we compute the FVD for videos generated with lengths over 400 frames, after training on the TaiChi dataset, and with over 1024 frames, after training on the Sky dataset, and the results are provided in Figures~\ref{fig:plot_fvd_longlen_vs}(a) and \ref{fig:plot_fvd_longlen_vs}(b), respectively. We compare the results provided by REncGAN3 to those of DIGAN~\cite{digan2022iclr} and TATS \cite{ge2022tatsLong}.  For DIGAN and TATS, we evaluate the generated video results provided by the TATS 
website\footnote{\url{https://songweige.github.io/projects/tats/index.html\#uncond-long}.}.
We can observe from Figure~\ref{fig:plot_fvd_longlen_vs}(a) that the generated videos by REncGAN3 with lengths over 400 frames on the TaiChi dataset have better FVD results than the videos generated by TATS \cite{ge2022tatsLong} and DIGAN \cite{digan2022iclr}. Meanwhile, Figure~\ref{fig:plot_fvd_longlen_vs}(b) shows that TATS and DIGAN have better FVD results on the Sky dataset. This happens because TATS and DIGAN aim to model short-term video information which is then repeated almost identically when creating long video sequences. Such repetitions do not influence much the video quality when changes are rather random, which is the case with the videos from the Sky dataset, but they would become visibly annoying in the context of videos showing structured movements, as in the TaiChi videos for example.

\subsubsection{The long-temporal video generation ablation study}

In this section, we consider variations of the loss function and in the number of overlapping frame numbers for the REncGAN3 when trained on the TaiChi dataset for generating long-term videos of $64\times 64$ pixel resolution.

\begin{table}[h]
   \centering
   \caption{FVD results of different video lengths when using only the overlapping video inputs (OVI) for the Encoders, merging the generated videos (MGV), or using both of these mechanisms (OVI+MGV) as in the Recall EncGAN3.}
   \label{tab_ablMarkov}
   \vspace*{0.4cm}
   \begin{tabular}{c|ccc}
       \hline
      Video Length  & OVI & MGV & Recall \\
       \hline \hline
       10  &   2254.30 &   322.91  &   199.98  \\
       16  &   2452.01 &   359.60  &   230.31  \\
       32  &   3052.51 &   340.69  &   283.20  \\
       64  &   2242.24 &   312.77  &   259.97  \\
       96  &   2505.06 &   288.66  &   243.44  \\
       100 &   2523.67 &   289.96  &   232.93  \\
       128 &   2717.66 &   326.78  &   252.51  \\
       136 &   2967.59 &   -      &   241.25  \\  
   \end{tabular}
\end{table}

{\bf Ablating the recall mechanism.} As mentioned in the beginning of Section~\ref{TrainRecall}, the Recall mechanism consists of two changes: presenting the Overlapped Video Input (OVI) clips to the two-stream encoder and Merging Generated Video (MGV) clips in the Discriminator. 
As illustrated in Figure~\ref{fig:REncGAN3}, the Markov chain in REncGAN3 is built by means of these two changes. 
The ablation results for FVD, when training on the TaiChi dataset, are provided in Table~\ref{tab_ablMarkov} which shows the results for generating videos of various lengths when considering only either OVI or MGV, as well as when both are used, as in the proposed Recall mechanism for generating long videos. The results from Table~\ref{tab_ablMarkov} show that both steps are needed by the REncGAN3 for generating long-term videos.

\begin{table}[h]
    \centering
    \caption{Ablating variations of loss functions with FVD of different video lengths.}
    \label{tab_ablLoss}
    \vspace*{0.4cm}
    \hspace*{-0.3cm}
    \begin{tabular}{c|ccccc}
    \hline
        Video & \multicolumn{5}{c}{Loss Functions}    \\  
        length & $L_1$ & $L_2$ & $L_3$ & $L_4$ & $L_5$ \\
        \hline \hline
        10  &   199.98  &   104.84  &   97.67   &   2525.70 &   67.81    \\
        16  &   230.31  &   111.87  &   107.54  &   2522.76 &   81.68    \\
        32  &   283.20  &   131.40  &   106.10  &   2737.70 &   83.54    \\
        64  &   259.97  &   114.16  &   125.99  &   3041.56 &   78.91   \\
        96  &   243.44  &   116.19  &   128.19  &   3206.97 &   88.42    \\
        100 &   232.93  &   118.36  &   125.92  &   3224.24 &   90.73    \\
        128 &   252.51  &   149.29  &   128.07  &   3299.62 &   101.26   \\
        136 &   241.25  &   138.17  &   140.56  &   3379.64 &   104.04   \\
    \end{tabular}
\end{table}

\begin{table}[t]
    \centering
     \caption{FID results  when applying REncGAN3 on short video generation.}
     \label{tab-FID}
     \vspace*{0.3cm}
    \begin{tabular}{l|ccccc}
        \hline
            & UvA & Weizmann & KTH & UCF101 \\
        \hline \hline
        EncGAN3       & 87.63 & 83.35 & 72.59 & {\bf 91.18} \\
        REncGAN3      & {\bf 73.14} & {\bf 70.91} & {\bf 66.97} & 95.87 \\
    \end{tabular}
    \footnotetext{Lower value is better.}
\end{table}
        
\begin{table*}[t]
    \centering
    \caption{Quantitative evaluation for IS and its components when training on the short video sequences. $\uparrow$ means the higher value is better while $\downarrow$ means the lower value is better.}
    \label{tab-IS}
    \vspace*{0.3cm}
    \begin{tabular}{l|ccc|c}
        \hline
            & IS$\uparrow$ & Inter-Entropy $\uparrow$ & Intra-Entropy $\downarrow$ & Dataset \\
        \hline \hline
                    & 571.29 & 6.499 & 0.151 & UvA \\
        EncGAN3    & 42.60 & 3.959 & 0.207 & Weizmann \\
                    & 50.48 & 4.812 & 0.891 & KTH \\
                    & 33.87 & 6.699 & 3.177 & UCF101 \\ \hline
                    & 87.007 & 4.656 & 0.190 & UvA \\
        REncGAN3     & 35.329 & 3.804 & 0.239 & Weizmann \\
                     & 11.477 & 4.087 & 1.647 & KTH \\
                    & 57.121 & 5.827 & 1.782 & UCF101 \\
    \end{tabular}
\end{table*}

{\bf Changing characteristics of the video generation and the loss function.}
The REncGAN3 model, introduced in Section~\ref{TrainRecall}, is used after being trained on the TaiChi sequence, for generating videos of various lengths, such as $T= \{ 10, 16, 32, 64, 96, 100, 128, 136 \}$. This ablation study uses FVD as the evaluation metric and in order to normalize the results we sample the videos (except for those which are smaller) into sets of 16 consecutive frames and then evaluate FVD on each of these video segments, as in the evaluations for the EncGAN3.
The results are provided in Table~\ref{tab_ablLoss} when considering different variations of the loss function, denoted as $L_1$, $L_2$, $L_3$, $L_4$ and $L_5$. We consider the loss function $L_5$ when two consecutive video clips are merged by training simultaneously the Encoder and Generator as in the equations~\eqref{LossREncG} and \eqref{Loss3LDV}, corresponding to the training of the REncGAN3 model.
Then we consider the loss functions $L_1$, $L_2$, $L_3$, $L_4$, where the Encoder and Generator are trained separately, as for the EncGAN3, which is the approach used for generating short video sequences.
The loss function $L_1$ considers the evaluation of the merged video-clips while $L_2$ evaluates separately each consecutive short video clips without testing them as being merged. Meanwhile, $L_3$ considers only $L_{D_I}$ stream from Eq.~\eqref{LossLDI} and not the video stream evaluation, while $L_4$ considers only $L_{D_V}$ stream from Eq.~\eqref{LossLDV} and not the image content evaluation, while generating long-term videos.
From Table~\ref{tab_ablLoss} the results for $L_3$ are similar to those for $L_2$, while $L_4$ leads to the worst results, indicating the necessity to use the image reconstruction error for the long-term video generation loss function. 
It can be observed from Table~\ref{tab_ablLoss} that $L_5$, adopted by REncGAN3, provides the best FVD results. 

{\bf Varying the number of overlapping frames when merging consecutive video clips.}
In these experiments, we vary the number of overlapping frames between two consecutive video clips which are merged by the replay network when generating videos of different lengths, after training on the TaiChi database. The results when considering $T_c - r \in \{ 0, 2, 4, 8 \}$ overlapping frames within initial video clips of 16 frames, when generating video sequences of various sizes are provided in Table~\ref{tab_ablOverlap}.
In this ablation study, we consider the loss function $L_1$ from those considered above and set the dimension of motion latent codes ${\bf z}_v$ as 10.  
The results from Table~\ref{tab_ablOverlap} show that overlapping 4 frames (25 \% of the entire video-clip) provides the best results for the TaiChi sequence.

\begin{table}[h]
    \centering
    \caption{FVD calculated for generated videos of various length T, when varying the number of overlapping frames, given by $T_c-r$, after training on the TaiChi database.}
    \label{tab_ablOverlap}
    \vspace*{0.3cm}
    \begin{tabular}{c|cccc}
        \hline        
        \multirow{2}*{T}  & \multicolumn{4}{c}{$T_c-r$} \\
                           & 8 & 4 & 2 & 0 \\
        \hline \hline
        10  &   199.98  &   128.48  &   176.52  &   322.91   \\
        16  &   230.31  &   132.05  &   194.68  &   359.60   \\
        32  &   283.20  &   149.14  &   198.30  &   340.69   \\
        64  &   259.97  &   141.13  &   184.13  &   312.77   \\
        96  &   243.44  &   158.43  &   174.86  &   288.66   \\
        100 &   232.93  &   153.67  &   171.94  &   289.96   \\
        128 &   252.51  &   162.58  &   193.88  &   326.78   \\
        136 &   241.25  &   162.76  &   -       &   -        \\
    \end{tabular}
\end{table}

\subsection{Comparative evaluation of long video generation models}

\begin{figure*}[h]
    \centering
        \begin{tabular}{c}
            \includegraphics[width=1\linewidth]{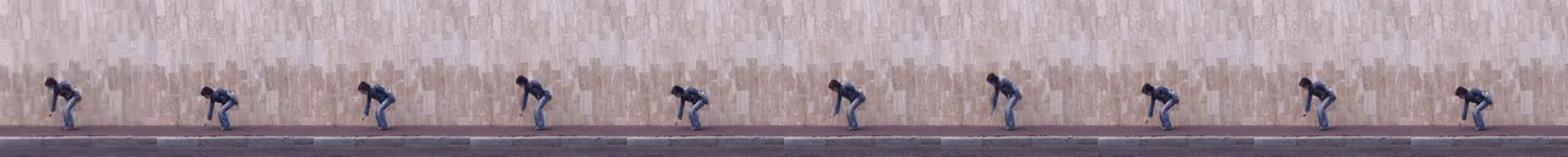} \\
            \includegraphics[width=1\linewidth]{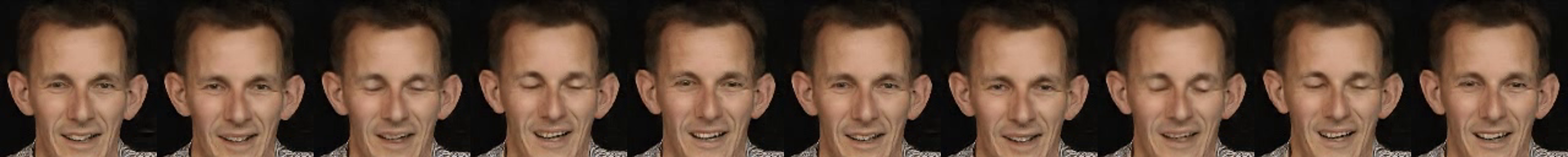} \\
            (a) REncGAN3 \\
            \includegraphics[width=1\linewidth]{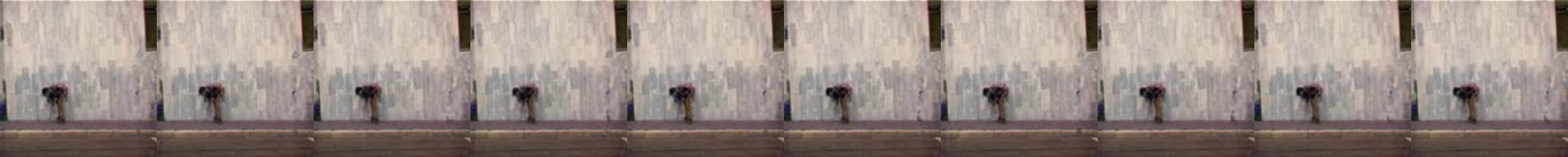} \\
            \includegraphics[width=1\linewidth]{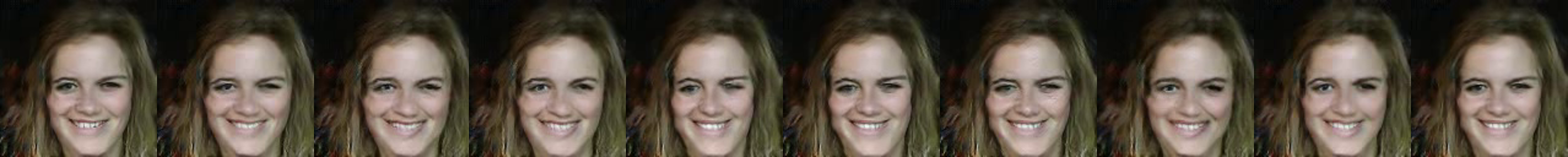} \\
            (b) EncGAN3 \\
        \end{tabular}
    \caption{Comparison between REncGAN3 and EncGAN3 when generating human action and facial expression videos, after training on the Weizzman (top row) and UvA (bottom row) databases. Each row shows 10 frames at $128\times128$ resolution, where one frame is sampled for each 5 frames from generated videos, covering in total a duration of 50 consecutive frames.}
    \label{fig:EncGvsR2_50f}
\end{figure*}

In this section, we compare the performance of REncGAN3 for long video generation with other methods. We first consider forcing EncGAN3 to generate longer videos, by increasing the size of the motion latent code ${\bf z}_v$. 
We also evaluate FID and IS 
when using the long-term video generation loss function from Eq.~\eqref{LossREncG}, by jointly training the Encoder and Generator, as in the REncGAN3, and then compare the results to those of the EncGAN3, where the two modules are trained separately. The results when considering UvA, Weizmann, KTH and UCF101 are provided in Tables~\ref{tab-FID} and \ref{tab-IS} for FID and IS, respectively. 
From these results, we can observe that by training the Encoder and Generator, as in REncGAN3 provides better results for most datasets when compared with EncGAN3.
Frames from videos, which are longer than 16 frames,  generated by REncGAN3 and EncGAN3 are provided in Figures~\ref{fig:EncGvsR2_50f}(a) and \ref{fig:EncGvsR2_50f}(b), respectively. The results indicate that REncGAN3 outperforms EncGAN3 in generating long-duration videos, particularly in maintaining high-quality details such as sharp human actions and clear facial expressions with realistic blinking.

\begin{figure*}[h]
    \centering
        \begin{tabular}{c}
            \includegraphics[width=1\linewidth]{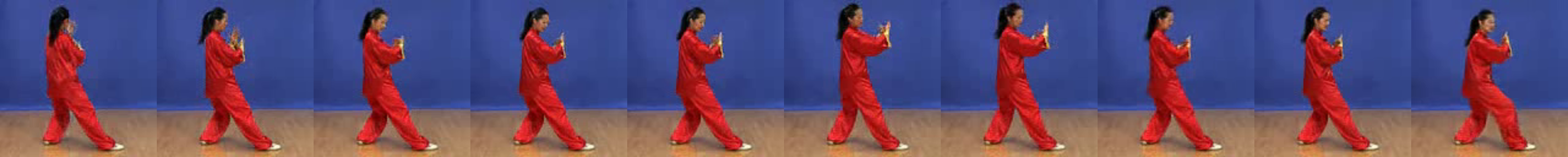} \\
            (a) REncGAN3 \\
            \includegraphics[width=1\linewidth]{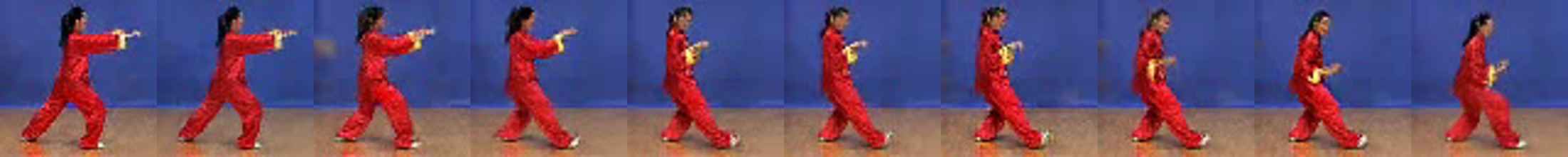} \\
            (b) LEncGAN3 \\
        \end{tabular}
    \caption{Video generation results by REncGAN3 and LEncGAN3  after training on the Taichi dataset. Frames in each row  are sampled per 30 frames each, from a generated video, covering a duration of 300 frames in total. Each generated frame has a resolution of $128\times128$ pixels.}
    \label{fig:LEGvsR2}
\end{figure*}

We also consider using a popular video modeling approach, the Long-Short-Term Memory (LSTM) \cite{hochreiter1997lstm,shi2015convlstm}, for extending temporally the video generation capabilities of EncGAN3, resulting in a method called LEncGAN3. Similarly to the REncGAN3, LEncGAN3 models the inter-clip relationship using a Markov chain framework, but it leverages LSTM to facilitate this modeling. LEncGAN3 applies LSTM modules at the ends of both content and motion encoders to process motion and content features before generating the latent spaces. The LSTM module inherits cell state information from the previous clip to learn the continuity information between consecutive clips.

Since most videos in the datasets used for training are of approximately 100 frames long, and many methods that simply increase the latent code size struggle to maintain good frame quality beyond this range \cite{digan2022iclr, wang2023styleinv}, the primary comparison between EncGAN3 and REncGAN3 focuses on generating videos of around 100 frames. However, we consider generating videos of more than 400 frames for REncGAN3 and LEncGAN3, given that both are designed specifically for long video generation. The results for REncGAN3 and LEncGAN3 are provided in Figures~\ref{fig:LEGvsR2}(a) and \ref{fig:LEGvsR2}(b) when considering training on the TaiChi database. From the frames shown in these figures, it can be observed that LEncGAN3 tends to produce artifacts in the background, whereas REncGAN3 consistently delivers higher-quality videos, particularly excelling in the generation of facial features and by smoothly modeling the movement of the hands. These results highlight the superior performance of REncGAN3 in long video generation tasks.

\section{Conclusions}
\label{sec-Con}

In this paper, we introduce a new video generation approach by enabling a GAN video generator with inference mechanisms provided by a variational encoder, resulting in a hybrid VAE-GAN video generating architecture. In line with other video processing architectures, the video generator consists of two generating streams for scene content and movement. The resulting Encoder GAN3 (EncGAN3) is shown to provide better videos than other models when generating short-term clips. Then we extend this approach for generating long-temporal video sequences by using a recall mechanism resulting in the Recall EncGAN3 (REncGAN3) which enforces the continuity between generated consecutive video segments by merging them and assessing their consistency within longer synthesized sequences. 
The efficiency of the proposed video generator in synthesizing video sequences of up to one minute, displaying continuous and consistent complex realistic movements is shown in the results following the training on several datasets. In future work the proposed recall mechanism will be applied  on the high quality short video sequences generated by other models, such as video diffusion generative models, for producing long video sequences.

\bibliographystyle{plain}
\bibliography{REncGAn3-arxiv}
\end{document}